\documentclass{article}
\usepackage[nonatbib, preprint]{neurips_2021}
\usepackage{url}
\usepackage{latexsym,amssymb}
\usepackage[cmex10]{amsmath}
\usepackage{soul}
\usepackage{tikz}
\usepackage{tikz-network}

\usepackage{caption}
\usepackage{subcaption}
\usepackage{multirow}
\usepackage{booktabs}
\usepackage{bm}
\usepackage{array}
\newcolumntype{C}[1]{>{\centering\let\newline\\\arraybackslash\hspace{0pt}}m{#1}}

\usepackage{pifont}

\newcommand\T{\rule{0pt}{2.9ex}}       
\newcommand\B{\rule[-1.2ex]{0pt}{0pt}} 

\title{Implications of Topological Imbalance for Representation Learning on Biomedical Knowledge Graphs}

\author{
    Stephen Bonner\textsuperscript{1}\footnotemark[1],\enskip Ufuk Kirik\textsuperscript{1}\thanks{Authors contributed equally to this work}, \enskip Ola Engkvist\textsuperscript{2},\enskip Jian Tang\textsuperscript{3,4,5},\enskip Ian P Barrett\textsuperscript{1} \\
\rule[10pt]{0pt}{0pt}
{\normalsize \textsuperscript{1}Data Sciences and Quantitative Biology, Discovery Sciences, R\&D, AstraZeneca, Cambridge, UK}\\
{\normalsize \textsuperscript{2}Molecular AI, Discovery Sciences, R\&D, AstraZeneca, Gothenburg, Sweden}\\
{\normalsize \textsuperscript{3}HEC Montreal, Canada} \\
{\normalsize \textsuperscript{4}Mila - Quebec AI Institute, Montreal, Canada}\\
{\normalsize \textsuperscript{5}Canadian Institute for Advanced Research (CIFAR), Canada}\\
\rule[15pt]{0pt}{0pt}
}

\begin{document}
\maketitle

\begin{abstract}

    Adoption of recently developed methods from machine learning has given rise to creation of drug-discovery knowledge graphs (KG) that utilize the interconnected nature of the domain. Graph-based modelling of the data, combined with KG embedding (KGE) methods, are promising as they provide a more intuitive representation and are suitable for inference tasks such as predicting missing links. One common application is to produce ranked lists of genes for a given disease, where the rank is based on the perceived likelihood of association between the gene and the disease. It is thus critical that these predictions are not only pertinent but also biologically meaningful. However, KGs can be biased either directly due to the underlying data sources that are integrated or due to modeling choices in the construction of the graph, one consequence of which is that certain entities can get topologically overrepresented. We demonstrate the effect of these inherent structural imbalances, resulting in densely-connected entities being highly ranked no matter the context. We provide support for this observation across different datasets, models as well as predictive tasks. Further, we present various graph perturbation experiments which yield more support to the observation that KGE models can be more influenced by the frequency of entities rather than any biological information encoded within the relations. Our results highlight the importance of data modeling choices, and emphasizes the need for practitioners to be mindful of these issues when interpreting model outputs and during KG composition.

\end{abstract}

\section{Introduction}\label{sec:intro}

Drug discovery and development is a highly complex, lengthy and multi-disciplinary endeavour~\cite{cook2014lessons}, associated with high costs and attrition, particularly due to efficacy failure~\cite{morgan2018impact} which is in turn underpinned by the hypothesis of which drug target(s) is thought to play a key role in a given disease. It is a process entailing many different critical decision points, for example which protein target(s) to focus on, what type of pharmacological agent to use to modulate a target, and which molecule to take into clinical trials. The biomedical domain is characterised by a high rate of technological evolution~\cite{goodwin2016coming, ran2013genome}, multitudes of low and high dimensional data types~\cite{ebrahim2016multi, hasin2017multi}, and a high volume and rate of semi-/unstructured information emerging constantly via scientific literature, conferences and patents.

These characteristics of the domain have led to exploration of Knowledge Graphs (KG) to improve representation of information, as well as delivery of predictions and hypotheses to improve the quality of the decisions made~\cite{bonner2021review, gaudelet2020utilising}. In a drug discovery KG, nodes often represent entities such as genes, diseases or drugs, whilst the relations between them capture their interactions with each other. Graph representation of these relationships not only allows for a more intuitive navigation of the complex domain for exploratory tasks, but is also particularly suitable for inferential analysis as Knowledge Graph Embedding (KGE) models can be used to derive meaningful, lower-dimensional representations of these biological entities and relationships. These representations in the embedding space can then be used for predicting missing links between these entities. In the case of target prediction, such approaches would produce a ranked list of genes, ordered by the score from the embedding model, predicted to be associated with a given disease. The top k elements from this list would then be inspected via subject matter experts (SME) to be triaged and further refined into a smaller list for a more manageable target validation process~\cite{paliwal2020preclinical}. For example, performing drug target identification has been addressed as link prediction between gene and disease entities using the ComplEx model~\cite{trouillon2016complex} on a drug discovery graph~\cite{paliwal2020preclinical}.

In the drug discovery domain, working with KGs has the promise of countering cognitive biases by leveraging massive amounts of information, well-beyond what is humanly feasible to comprehend, to inform a given hypothesis or decision through a computationally sound process that is agnostic to \emph{a priori} experience. Challenges remain however, e.g.\ dealing effectively with the structural biases in the information sources~\cite{nguyen2017pharos} and the relative lack of negative data for validation in particular. To the best of our knowledge, the precise effect of the structural imbalance inherent in biomedical KGs, on predictive tasks using KGE models has not been explored thoroughly in the literature.

In this study, we aim to fill in this gap by providing a comprehensive analysis of how structural imbalance in the underlying data in a KG affects the scores for predicted relationships. We demonstrate that the KGE methods tend to overrate the highly-connected entities, and that this result is reproducible in different datasets (i.e. graphs), for different embedding models, across different diseases and different predictive tasks. We go on to study the extent of this imbalance by further investigating the type of relationships the entities have, by introducing various types of perturbations to the graph to assess the stability of the rankings, as well as presenting a thorough case study in target discovery task set on the unperturbed graph. Seen altogether, our results would indicate that the total volume of connections an entity has within the graph seemingly matters more than any biological information encoded within. We thus highlight the need for careful consideration of graph composition in relation to the analytical methods employed for a given task.

Example code for the experiments presented in thus study is available online.\footnote{\url{https://github.com/AstraZeneca/biomedical-kg-topological-imbalance}} Results are presented on only publicly available datasets.
\section{Background \& Related Work}\label{sec:litreview}

\subsection{Background}\label{ssec:background}

\textbf{Knowledge Graphs.} A knowledge graph is a way to capture and represent, often widely scoped, knowledge within a domain, based on graph theory principles. Semantically, a KG is a heterogeneous, multi-relation and directed graph, connecting a set of entities \(\mathcal{E}\) to each other by a set of relationships \(\mathcal{R}\), defined as \(\mathcal{K} \subseteq \mathcal{E} \times \mathcal{R} \times \mathcal{E}\) \cite{zhang2019heterogeneous}. KGs are often considered as a series of triples \((h,r,t) \in \mathcal{K}\), where \( h,t \in \mathcal{E}\) are the head and tail entities connected via the relationship \( r \in \mathcal{R}\). A hypothetical triple from a drug discovery KG could be \(( \mathit{Gene}_i, \mathit{associates}, \mathit{Disease}_j) \), where the entities \(\mathit{Gene}_i\) and \(\mathit{Disease}_j\) are connected via the relationship \(\mathit{associates}\). Given that a KG is still a graph, its topological structure can be measured using a well studied range of techniques~\cite{newman2018networks}. A common measure to consider at the level of individual entities is that of \emph{degree}, which is the number of relationships of any type a given entity has to all others (possibly including itself) within the graph.

In many real-world KGs, the set of triples can be noisy and incomplete~\cite{ali2020bringing}. Thus numerous techniques attempt to impute missing information based on the existing data in \(\mathcal{K}\) through multi-relation link prediction~\cite{ruffinelli2019you}. Such techniques consider the partial triple \(( \mathit{Gene}_i, \mathit{associates}, ?) \) and attempt to predict the correct tail entity, i.e. try to answer the question \emph{`with which disease(s) is \(\mathit{Gene_i}\) associated?'} or be given \(( ?, \mathit{associates}, \mathit{Disease}_j) \) and attempt to predict the correct head entity, i.e. \emph{`which gene(s) are associated with \(\mathit{Disease_j}\)?'}

\textbf{Knowledge Graph Embeddings.} In this work we consider KG embedding models~\cite{ji2020survey, wang2017knowledge}, which aim to learn low-dimensional representations of all entities and relations. These embeddings are combined in various ways to produce a score of how likely a given triple is to be true, with a larger score typically implying a more plausible triple~\cite{ali2020bringing}. More concretely, a model \(f : \mathcal{E} \times \mathcal{R} \times \mathcal{E} \rightarrow \mathbb{R} \), calculates a scalar value \(s_{(h,r,t)}\) representing the plausibility for each potential triple \((h,r,t) \in \mathcal{K}\). For KGE approaches, \(f\) is a learned model which operates only with the embeddings of the elements in the triples, \(f( \mathbf{h}, \mathbf{r} , \mathbf{t} )\), where \(\mathbf{h}, \mathbf{t} \in \mathbb{R}^m \) and \(\mathbf{r} \in \mathbb{R}^n \). The values of \(m\) and \(n\) represent the dimension of the entity and relation embedding respectively. It should be noted that the values of the scaler scores \(s_{(h,r,t)}\) are not directly meaningful by themselves and are generally used to rank true triples above a sampled set of negative ones. Furthermore, the scale of score values varies according to the function used by the particular KGE model~\cite{ali2020bringing}, meaning comparing scores from different models side by side is meaningless.

Given a trained KGE model, it is possible to pass it an incomplete triple \((h,r,?)\) as described earlier and it will predict which entity is most likely to complete it. In practice, this is achieved by enumerating over all entities in the graph and the model scoring the triple with each entity taking the place of the missing element: \( \mathcal{S} = \{f(h,r,t^\prime) \mid t^\prime \in \mathcal{E} \} \). This list of scores \(\mathcal{S}\) is then sorted and a user can interpret it as a ranked list from most to least likely. Many of the common metrics, such as Hits@k or Mean Reciprocal Rank (MRR) for assessing the predictive performance of KGE models are based on this ranked output, with the hope that the model ranks true, but previously unknown triples, above meaningless ones~\cite{ali2020bringing}.

One of the most attractive properties of KGE methods is that it is possible to query the same trained model for link prediction tasks no matter the exact context as the model is trained to reconstruct the entire graph, agnostic of a particular type of inference. In other words, it is possible to reuse the same model across different tasks, whether predicting the most likely genes to associate with a given disease, or the most likely drugs to interact with a given gene.

\textbf{Drug Discovery and Knowledge Graphs.} Drug discovery is driven by the need to improve the standard of care for a particular disease, either by providing a more efficacious treatment or one that has fewer, less severe side effects. A crucial step early in the process is target discovery, that is to correctly identify molecular entities that are associated with the disease or symptom in question, for which a drug can be designed and developed. These entities can be genes, proteins or metabolites, which play an important role in disease causation or progression and alternatively play an important role in managing the symptoms of the disease (e.g. while steroid nose sprays like mometasone do not affect the direct cause of allergy, they help alleviate symptoms like running nose). With the cost of drug discovery estimated at hundreds of millions of dollars~\cite{wouters2020drugdisco}, it is crucial to find the right target~\cite{morgan2018impact}.

One use case for KG predictions is for them to be integrated into a complex drug discovery pipeline leading to resource intensive experiments being performed to validate the predictions~\cite{paliwal2020preclinical}. In the case of target prediction, such approaches would produce a ranked list of genes, ordered by the score from the model, predicted to be associated with a given disease. The top k elements from this list would then be inspected via subject matter experts (SME) to be triaged and further refined into a smaller list for a more manageable target validation process. In one study, the top 600 genes predicted to be associated with rheumatoid arthritis were reduced down to 55 by SMEs to be taken forward for experimental validation~\cite{paliwal2020preclinical}. It is therefore imperative that predictions are made using pertinent biological information so that viable targets can be discovered.

\subsection{Previous Works}

It has long been observed that graphs representing real-world data possess a long or heavy-tailed distribution of degree values~\cite{barabasi1999emergence}. This means that the majority of vertices have a limited number of connections to others, with a small number of highly connected hub-vertices being present. This inherent connectivity imbalance present in KG data can pose challenges when using it for input to machine learning models, however to what extent it has an impact has not yet been fully explored.

The issue of biased training data has become an area of great interest within the KG field. Typically, the bias of concern is that of attributes associated with the entities in the graph, for example gender, age or race~\cite{wang2021unbiased, arduini2020adversarial, bourli2020bias}. However, recent work~\cite{mohamed2020popularity} has shown how popularity bias is present in three frequently used non-biomedical KGs: FB15K~\cite{bordes2013translating}, WN18~\cite{bordes2013translating} and YAGO3-10~\cite{mahdisoltani2014yago3}. The work argues that traditional methods for addressing data imbalance, such as up-weighting or over-sampling minority classes, do not map well onto KG data where we operate in the unit of triples. By definition triples contain two linked entities, each of which might have a very different level of connectivity within the graph, so how to re-weight the training process of KG-specific models is unclear~\cite{mohamed2020popularity}. Instead, the authors propose alterations to the Hits@k and MRR ranking metrics to account for entity and relation popularity when scoring model performance.

The issue of non-uniform graph connectivity (typically in homogenous graphs) has begun to be studied in parallel by the field of Graph Neural Networks (GNN), where researchers have shown that models learn low-quality representations, and thus making more incorrect predictions for low-degree vertices~\cite{liu2020towards, liu2021tail, tang2020investigating}. This has also been explored in the context of homogenous graph representation learning~\cite{arduini2020adversarial} and for random walks~\cite{kojaku2021residual2vec, rahman2019fairwalk}.

Other work has argued that degree can be a strong feature for predicting an edge between two entities in biomedical graphs, but this can cause issues when the graph representation does not truly reflect the real underlying connections between the entities~\cite{zietz2020probability}. The work derives a prior via a process of graph perturbations to measure the likelihood of an edge being present based solely on degree. However the impact of connectivity in relation to KGE models is not considered.
\section{Experimental Setup}\label{sec:expsetup}

\subsection{Dataset}\label{ssec:dataset}

\begin{figure*}[!ht]
	\centering
	\begin{subfigure}[b]{0.3\textwidth}
		\centering
		\includegraphics[width=0.99\textwidth]{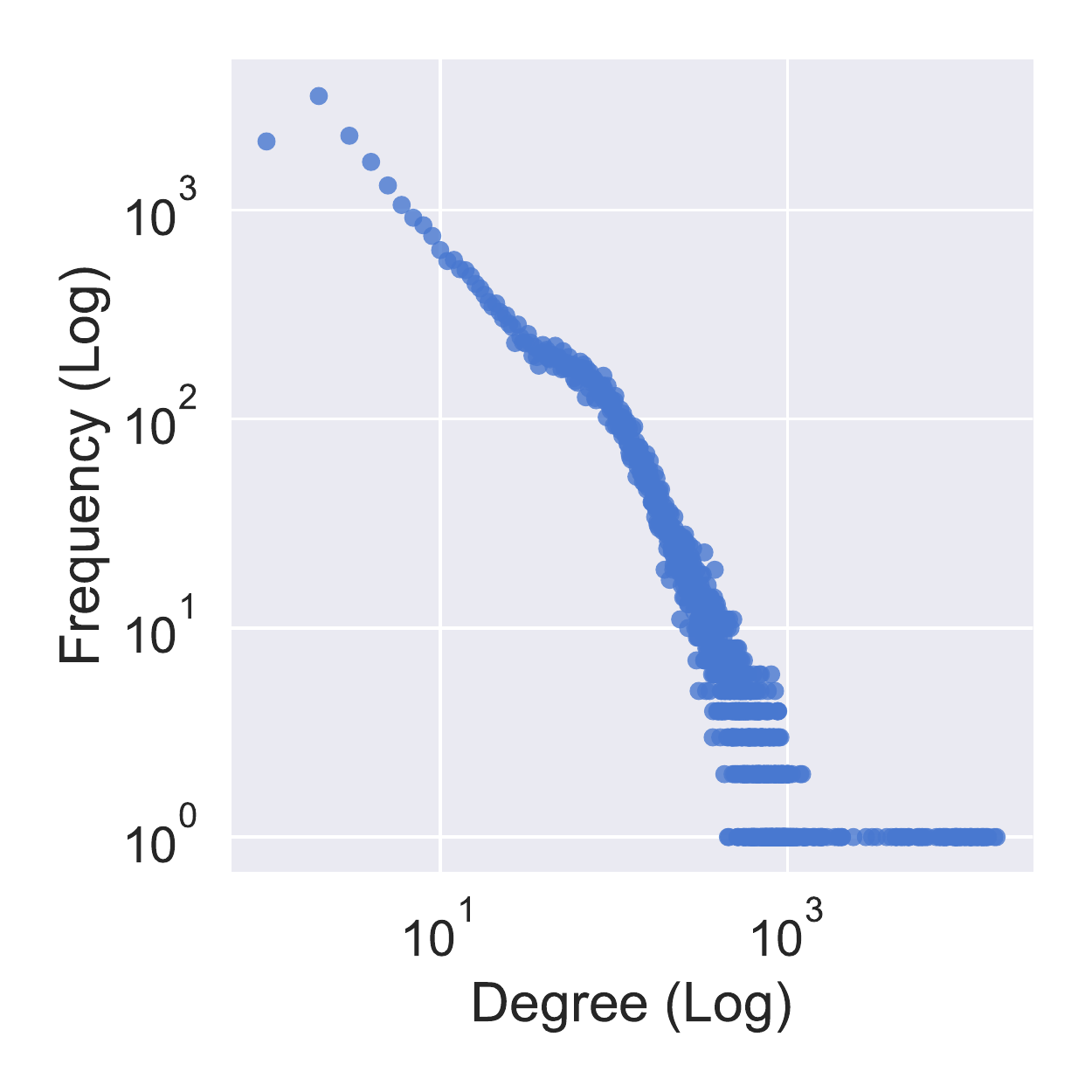}
		\caption{Degree Distribution}\label{fig:hetionet:full}
	\end{subfigure}
	\begin{subfigure}[b]{0.3\textwidth}
		\centering
		\includegraphics[width=0.99\textwidth]{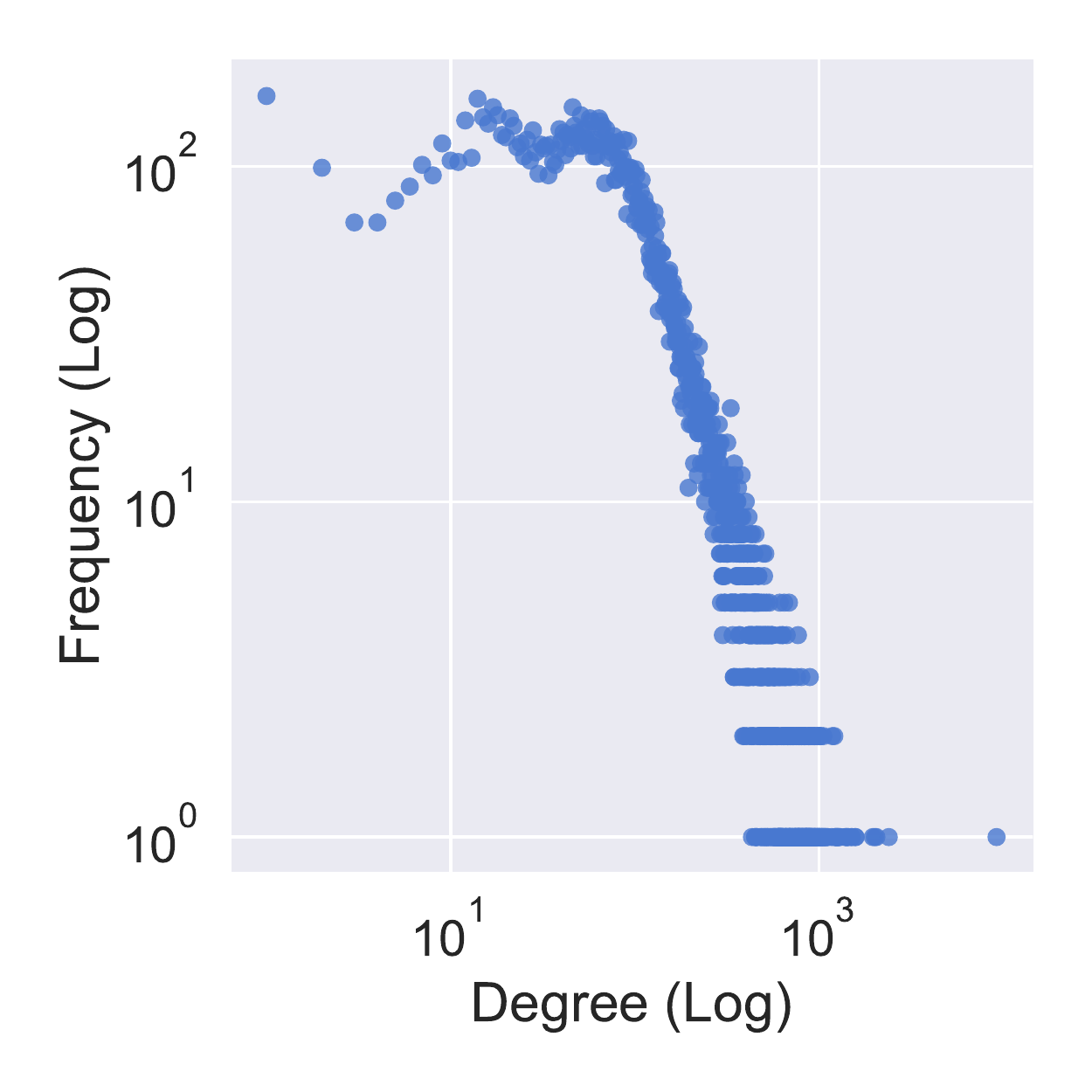}
		\caption{Degree Distribution (Gene)}\label{fig:hetionet:gene}
	\end{subfigure}
	\begin{subfigure}[b]{0.365\textwidth}
		\centering
		\includegraphics[width=0.99\textwidth]{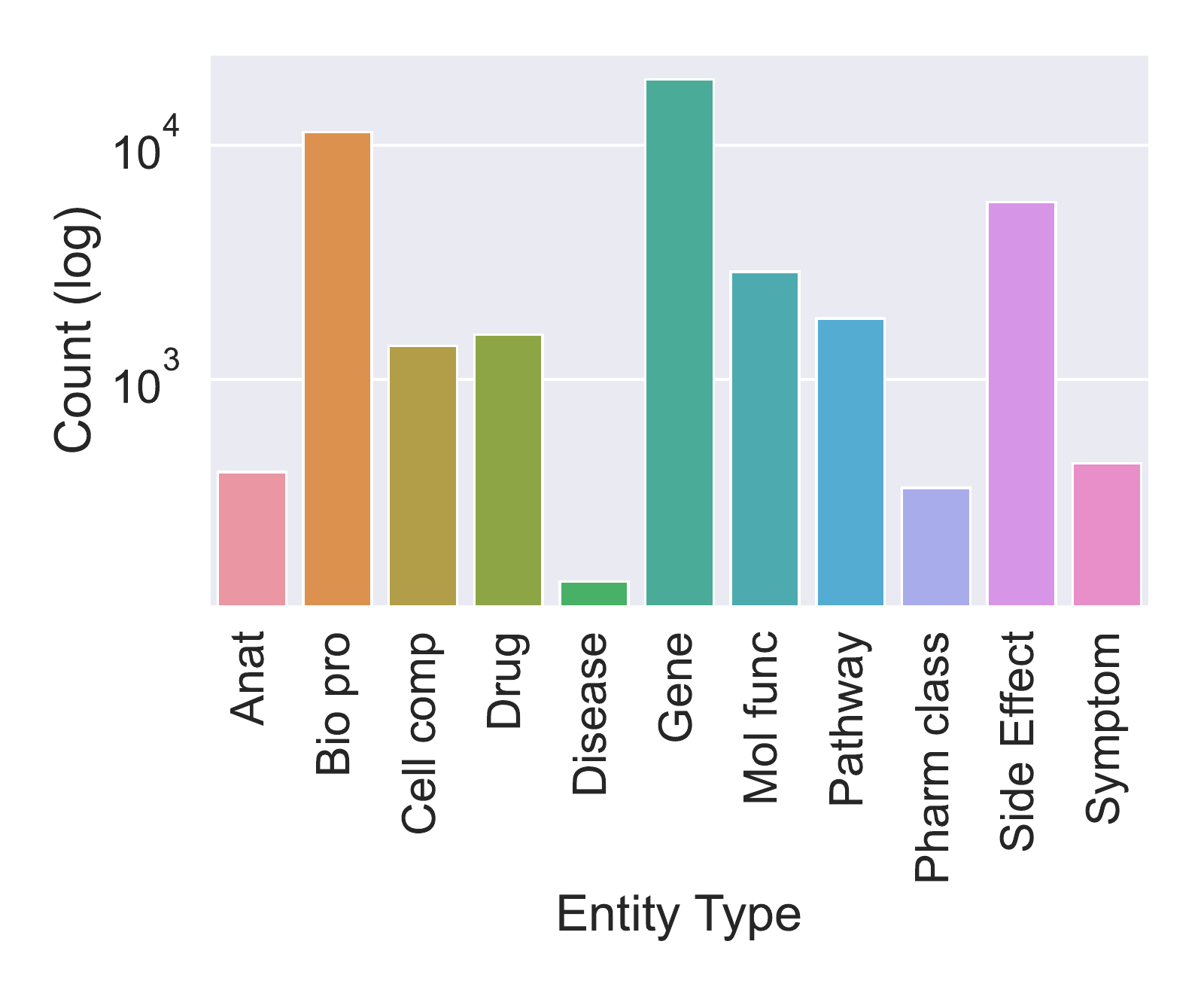}
		\caption{Entity Frequency}\label{fig:hetionet:ef}
	\end{subfigure}

	\begin{subfigure}[b]{0.5\textwidth}
		\centering
		\includegraphics[width=0.99\textwidth]{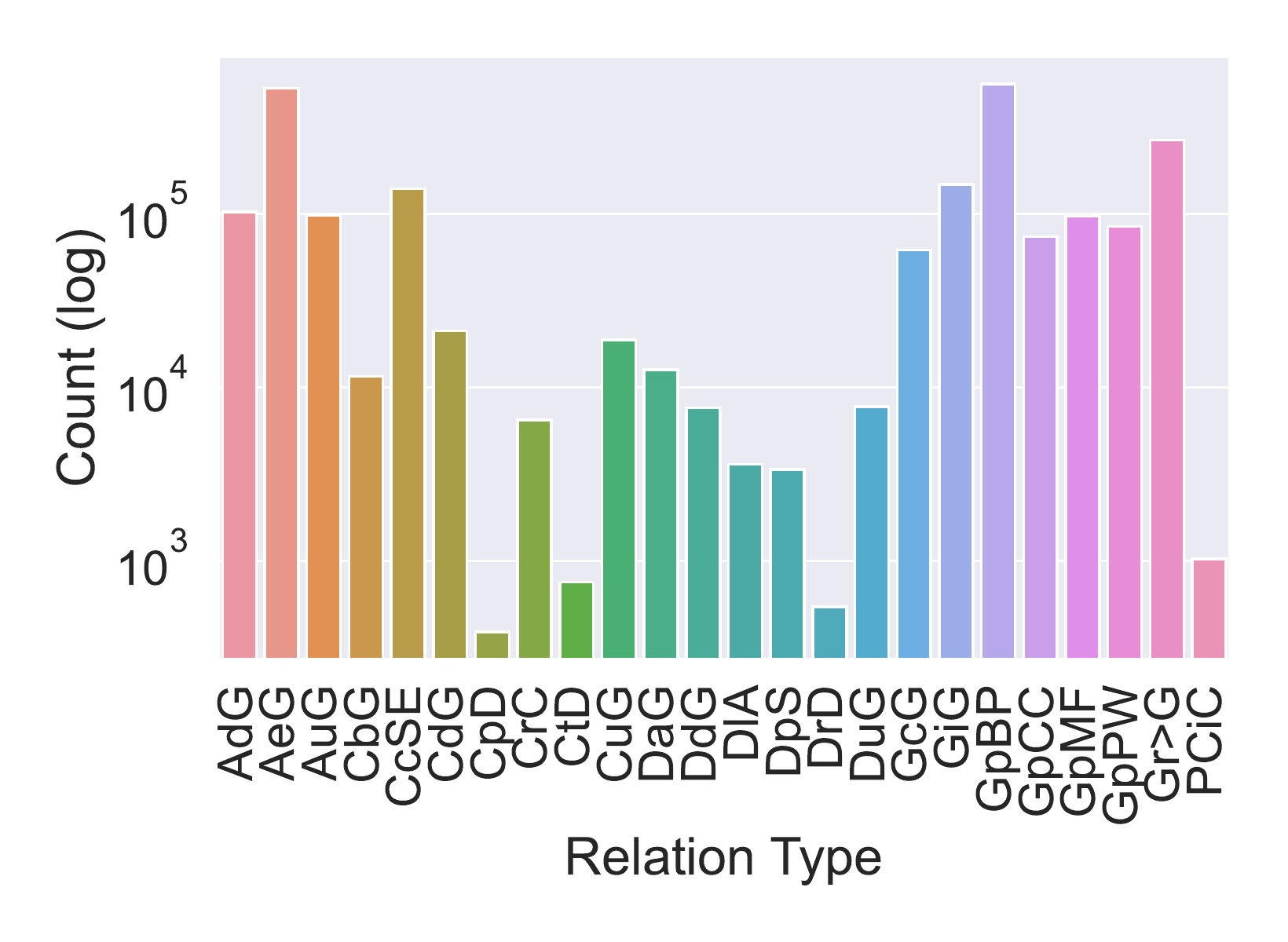}
		\caption{Relation Frequency}\label{fig:hetionet:rf}
	\end{subfigure}
	\caption{An overview of the Hetionet dataset.}
	\label{fig:hetionet}
\end{figure*}

Throughout this work we primarily make use of the publicly available Hetionet biomedical knowledge graph \cite{himmelstein2017systematic}. Hetionet was originally created as part of a project focusing on drug repurposing and integrates 29 public data sources, however its use has been explored in other areas such as target prediction~\cite{bonner2021understanding}. Hetionet contains information on diseases, human protein-coding genes and compounds, among others, which are represented as entities within the graph and total over 47K. These are linked via 24 different relationship types capturing 2.2M interactions between them. An overview of Hetionet is presented in Figure \ref{fig:hetionet}, which shows the overall degree distribution (Figure \ref{fig:hetionet:full}), as well as for just the gene entities (Figure \ref{fig:hetionet:gene}), both of which demonstrate a heavy-tailed distribution. The frequency of the entities (Figure \ref{fig:hetionet:ef}) and relations (Figure \ref{fig:hetionet:rf}) shows an imbalance, with certain ones being more prevalent than others\footnote{A detailed explanation of entity and relation types can be found in the original publication~\cite{himmelstein2017systematic}}.

As we focus on target discovery, that being the prediction of links between gene and disease entities, we make use of the \(\mathit{Disease} \xrightarrow[]{\textit{associates}} \mathit{Gene}\) (DaG) edges from Hetionet. These edges are extracted from data sources such as DISEASES~\cite{pletscher2015diseases}, GWAS Catalog~\cite{buniello2019nhgri} and DisGeNET~\cite{pinero2016disgenet}, all of which capture known gene-disease associations and come from a variety of sources, ranging from expert curation to text mining. Additionally, Hetionet also contains two other edge types between gene and disease entities: \(\mathit{Disease} \xrightarrow[]{\textit{upregulates}} \mathit{Gene}\) and \(\mathit{Disease} \xrightarrow[]{\textit{downregulates}} \mathit{Gene}\). This is important to consider as it could be trivial for a model to learn that if a gene-disease pair are linked via an up- or down-regulates edge, then they should also be linked via an associates edge, as it would be a common case logically.

\subsection{Model}\label{sec:model}

Many knowledge graph embedding models have been introduced, with the primary differentiator between them being how they score the plausibility of a given triple. Throughout this present study we make use of the TransE model which was one of the earliest approaches, and often still one of the most capable. TransE uses the notion of translational distance to learn embeddings such that relations are used to translate between entities in latent space~\cite{bordes2013translating}. In the TransE score function, the relation embedding is added to the head entity such that the result lies close to the tail embedding:
\[ f(h,r,t) =  -|| \mathbf{h} + \mathbf{r} - \mathbf{t}||_F, \]
where \(F\) can be the l1 or l2 norm. One can interpret this score as simply the distance between the head and tail entity in the high-dimensional embedding space. Since its introduction, numerous other approaches have been proposed in the literature which build on or alter the TransE approach including TransH~\cite{wang2014knowledge}, ComplEx~\cite{trouillon2016complex}, RotatE~\cite{sun2019rotate} \& DistMult~\cite{yang2015embedding}, all with different strengths and weaknesses regarding relation types that can be captured~\cite{rossi2021knowledge}. However TransE has proven to still be highly competitive, when tuned appropriately, and can outperform more recent approaches~\cite{bonner2021understanding}.

Considering its comparable predictive performance and relatively modest computational requirements, the majority of our results refer to TransE, as the experiments we demonstrate in this study required a large number of training regimens. Nevertheless, we train and evaluate the performance of the above-mentioned newer models in the gene-disease prediction setting, and indeed observe the same trends regardless of the model in use (See Figure \ref{fig:degree-regression-models}). Furthermore, all these models were trained using optimal hyperparameters previously established~\cite{bonner2021understanding} on the same dataset.

\subsection{Case Study Diseases}\label{ssec:dis}

Not all entities are created equally and in our context this refers to the connectivity of diseases which naturally varies due to how well-studied a particular disease is. In order to assess how the ranking of \(DaG\) predictions vary due to the degree of the disease, we choose diseases on both ends of the connectivity scale. An overview of these five diseases and how they are connected within the graph is presented in Table \ref{tab:disease}.

On the upper end, we have Breast Cancer which is the most highly-connected disease in Hetionet, with a degree of 1159, out of which 540 are of type \(DaG\). However, considering the heterogeneous nature of breast cancers and the diversity of both clinical and molecular subtypes~\cite{szymiczek2021bc} we also wanted to include other highly-connected diseases, with relatively more homogeneous aetiologies in the analysis. For this purpose, we chose melanoma and Parkinson's disease (PD) as representatives of  highly-studied diseases. Melanoma is a common type of cancer with relatively few subtypes and many known driver mutations \cite{rabbie2019melanoma}. Furthermore, it is the 4th best connected disease in the Hetionet graph, with a degree of 944 and 930 distinct neighbours. Parkinson's, on the other hand, is a complex neurodegenerative disease that has several monogenic variants, that is mutations in single genes causing the disease, as well as various identified genetic risk factors~\cite{hernandez2016Parkinson}. In Hetionet, Parkinson's disease has a degree of 795, and is the 8th most connected disease.

For low connectivity diseases, we chose Fuchs Endothelial Dystrophy (FED) and Fallopian Tube Cancer (FTC). Fuchs endothelial dystrophy, is a relatively rare disease with only 22 connections in the graph, all distinct neighbors. Besides being the 4th least connected disease in the graph, we found this disease to be of interest in terms of high level of characterisation and several known genetic factors as well as affected bioprocesses~\cite{Eghrari2015Fuchs}. Fallopian tube cancer is a fairly rare type of cancer, with an incidence rate of about 0.36-0.41 per 100000 women~\cite{stasenko2019ftc}. Given its rarity, there is little in literature about this disease, especially in comparison to closely related ovarian and peritoneal carcinomas. In Hetionet, FTC has a degree of 29; connecting to 10 diseases, 8 symptom and anatomy nodes each, and most importantly to 3 genes.

Seen together these diseases provide a wide range of pathologies, both common and rare. Additionally, these diseases have high tissue-specificity and, in most cases, relatively low diversity in cell types and subtypes. Even in the case of FED and FTC, there are known genes with causative and/or risk-affecting mutations. These properties can be used in order to sanity check the predicted gene-disease associations to some extent.

\begin{table*}[ht!]

	\centering
	\begin{tabular}{p{0.295\textwidth}  C{0.1\textwidth}  C{0.15\textwidth} C{0.15\textwidth} C{0.15\textwidth}}
		\toprule
		\textbf{Disease}            & \textbf{Degree} & \textbf{\# Distinct} & \textbf{\# DaG} & \textbf{Connectivity} \\
		\midrule \midrule
		Breast Cancer               & 1159            & 1123                 & 540             & High                  \\
		Melanoma                    & 944             & 930                  & 342             & High                  \\
		Parkinson's Disease         & 795             & 789                  & 143             & High                  \\
		\midrule
		Fallopian Tube Cancer       & 29              & 29                   & 3               & Low                   \\
		Fuchs Endothelial Dystrophy & 22              & 22                   & 6               & Low                   \\
		\bottomrule
	\end{tabular}
	\vspace{5pt}
	\caption{The diseases we focus on in this study and associated metrics for them from Hetionet. Where \# Distinct is the number of unique entities connected to the disease.}
	\label{tab:disease}
\end{table*}

\subsection{Implementation Details}

All work has been performed using the open-source PyKEEN framework~\cite{ali2020pykeen}, a python library for knowledge graph embeddings built on top of PyTorch~\cite{paszke2019pytorch}. All experiments were performed on machines with Intel(R) Xeon(R) Gold 5218 CPUs and NVIDIA(R) V100 32GB GPUs. Additionally, we kept the software environment consistent throughout all experimentation using python 3.8, CUDA 10.1, PyTorch 1.9, and PyKEEN 1.5.0. Models were trained using optimal hyper-parameters, presented in Table~\ref{tab:params}, as discovered through a detailed optimisation process of over 100 unique parameter configurations per model~\cite{bonner2021understanding}. Additionally, all models were trained on the same fixed random split of the Hetionet dataset.

\begin{table}[h!]
	\centering
	\small
	\begin{tabular}{l c c c c c }
		\toprule
		\textbf{Parameter} & \multicolumn{5}{c}{\textbf{Value By Approach}} \T\B                                                                      \\
		\midrule \midrule
		                   & ComplEx                                                                          & DistMult & RotatE & TransE & TransH\B \\
		\cline{2-6}

		Embedding Dim      & 272                                                                              & 80       & 512    & 304    & 480\T    \\
		Num Epochs         & 700                                                                              & 400      & 500    & 500    & 800      \\
		Learning Rate      & 0.03                                                                             & 0.02     & 0.03   & 0.02   & 0.005    \\
		Num Negatives      & 91                                                                               & 41       & 41     & 61     & 1        \\
		\midrule
		Optimiser          & \multicolumn{5}{c}{Adagrad}                                                                                              \\
		Inverse Relations  & \multicolumn{5}{c}{False}                                                                                                \\
		Loss Function      & \multicolumn{5}{c}{Negative Sampling Self-Adversarial Loss~\cite{sun2019rotate}}                                         \\
		\bottomrule
	\end{tabular}
	\caption{Model hyperparameters and training setups.}\label{tab:params}
\end{table}

\emph{A note on runtime:} The training of knowledge graph embedding models can be costly in regards to runtime on KGs the size of Hetionet~\cite{bonner2021understanding}. For example, training a TransE model with the hyperparameters detailed in Table~\ref{tab:params}, took six hours on the previously detailed hardware. All other models demonstrated runtimes equal to, or often, greater than this. This has particular implications for the experiments where the KG is being perturbed in some way (i.e.\ the edge removal, addition or rewiring experiments) as each individual data point and each experiment repeat requires a full model retraining, meaning that these results took several hundred GPU hours to compute. This is ultimately why we limited this analysis to specific diseases, rather than repeating it over all available in Hetionet.
\section{Results}\label{sec:results}

In this section we present the results of our experimental evaluation. Unless otherwise stated, all experiments employ the \(\mathit{Disease} \xrightarrow[]{\mathit{associates}} \mathit{Gene}\) relation type from Hetionet and make use of the TransE model as described in Section~\ref{sec:expsetup}.

\subsection{Correlation with Topological Structure}\label{ssec:cor-top}

We begin by studying correlation between an entity's degree and the score assigned to it by the model measuring how likely it is to be the correct one to complete the triple.

\begin{figure*}[!th]
	\centering
	\begin{subfigure}[b]{0.48\textwidth}
		\centering
		\includegraphics[width=0.99\textwidth]{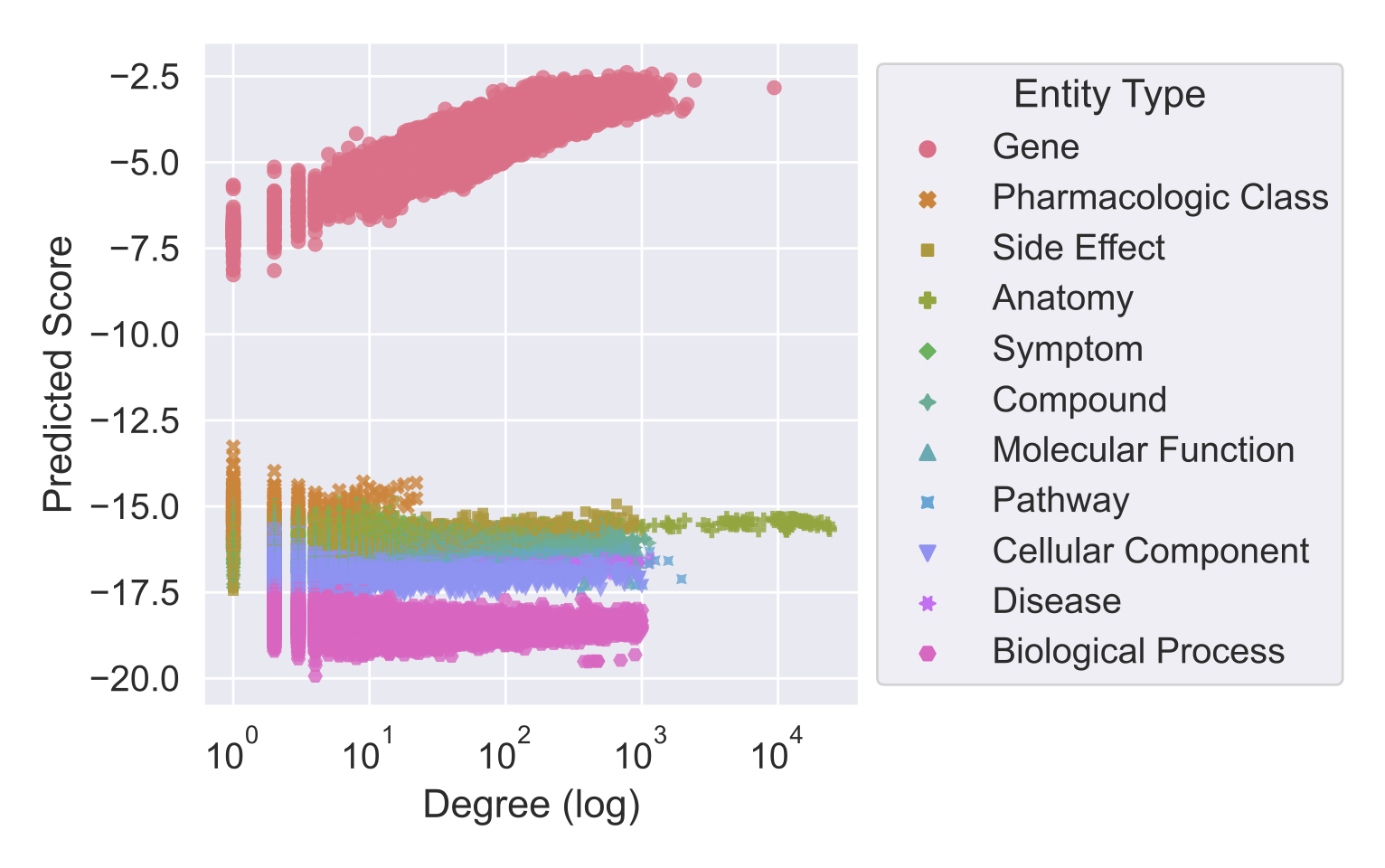}
		\caption{Melanoma}\label{fig:ae:mel}
	\end{subfigure}
	\begin{subfigure}[b]{0.48\textwidth}
		\centering
		\includegraphics[width=0.99\textwidth]{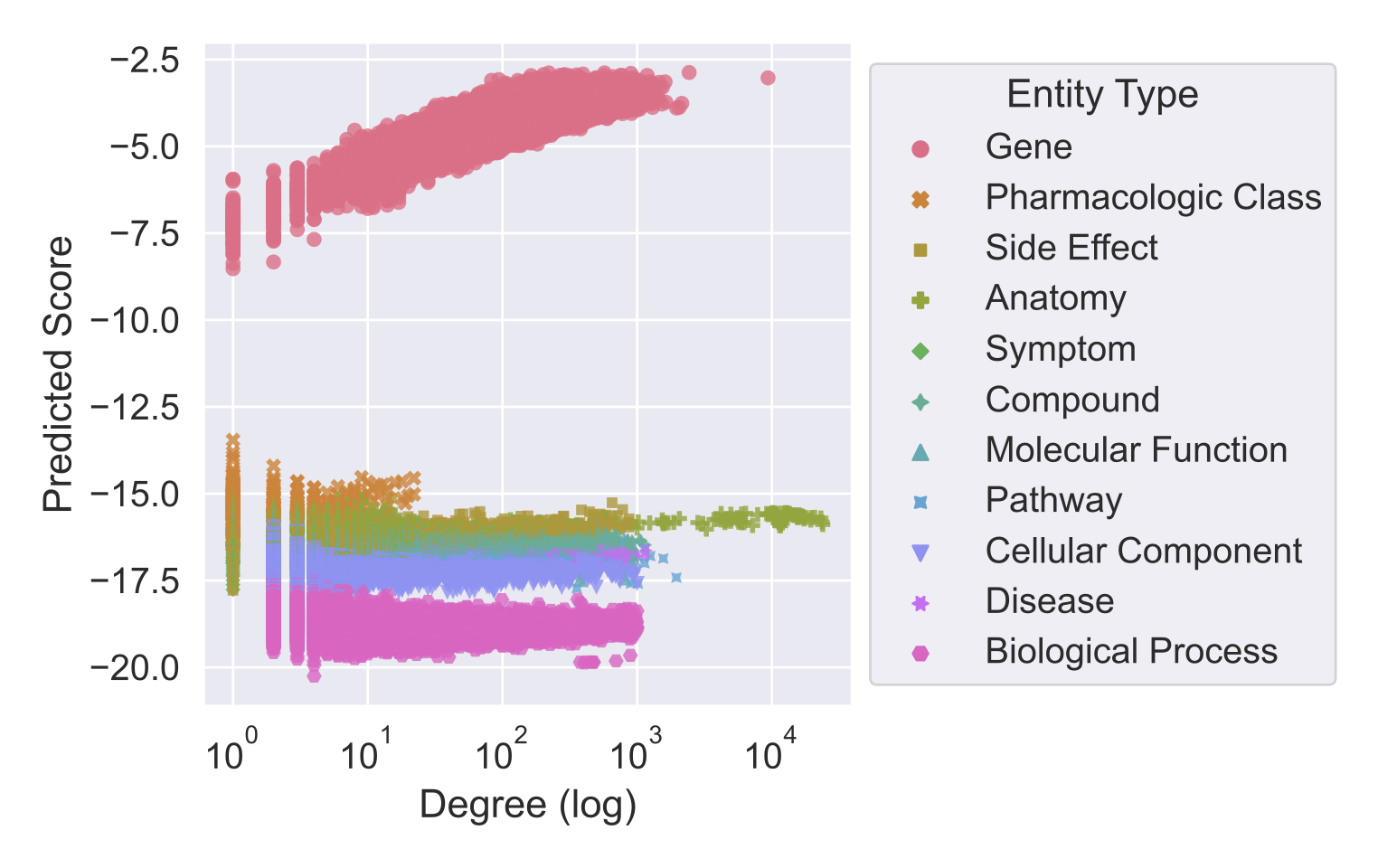}
		\caption{Fuchs Endothelial Dystrophy}\label{fig:ae:fuchs}
	\end{subfigure}
	\caption{Predicted score for each entity using TransE when completing the partial triple (Disease, DaG, ?) versus the degree of the entity. Points are styled by entity type. The higher the score, the more likely the model considers the entity and being the correct one to complete the triple.}
	\label{fig:all-entity}
\end{figure*}

As highlighted in Section \ref{ssec:background}, KGE models are essentially agnostic of the data schema when completing missing edges during inference time. As such they provide a score for each and every entry regardless of their type. Figure \ref{fig:all-entity} displays the score assigned by TransE\footnote{This score is the direct value taken from the TransE scoring function as detailed in Section \ref{sec:model}, hence the negative value.} to each entity in the Hetionet graph when completing the triple (Disease, DaG, ?) plotted against the degree for that entity, from which two main patterns can be seen. First and foremost, in both Melanoma (Figure \ref{fig:ae:mel}) and FED (Figure \ref{fig:ae:fuchs}) we can see a clear separation for Gene entities from everything else in the score space. This implies that the model is assigning a higher confidence to Genes being the correct entities to complete the triple. This is encouraging since we can conclude that the model at least is able to learn that the DaG relation only ever connects a Disease and a Gene in the training dataset.

Having passed this first sanity check for the predictions, we thus move to focus on the relationship between the score assigned to just the Gene entities. The second pattern observed in Figure \ref{fig:all-entity} was that there appears to be a correlation between degree and score. We now explore this in further detail for the various diseases explained in Section~\ref{ssec:dis}.

\begin{figure*}[!ht]
	\centering
	\begin{subfigure}[b]{0.48\textwidth}
		\centering
		\includegraphics[width=0.99\textwidth]{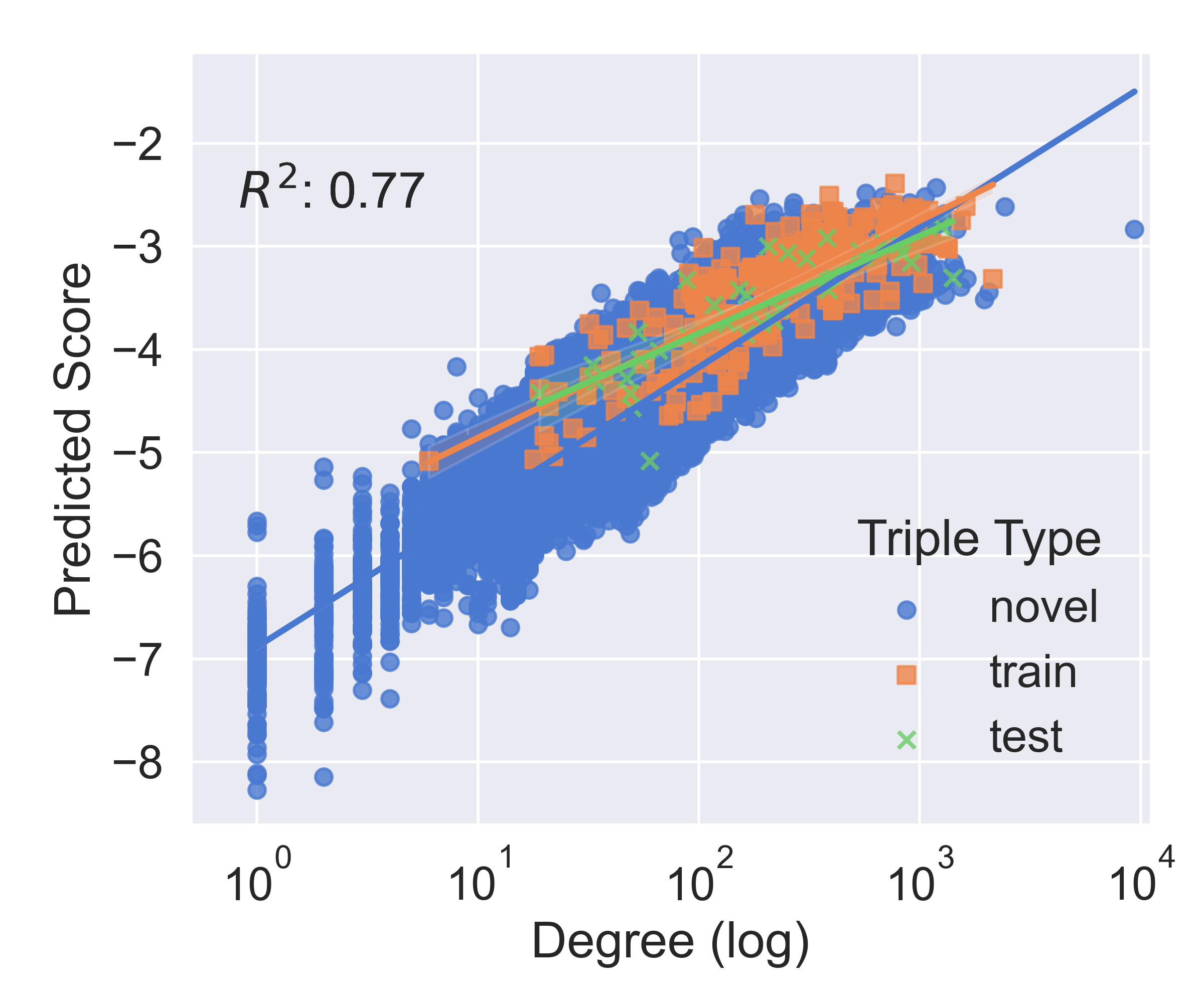}
		\caption{Melanoma}\label{fig:dr:mel}
	\end{subfigure}
	\begin{subfigure}[b]{0.48\textwidth}
		\centering
		\includegraphics[width=0.99\textwidth]{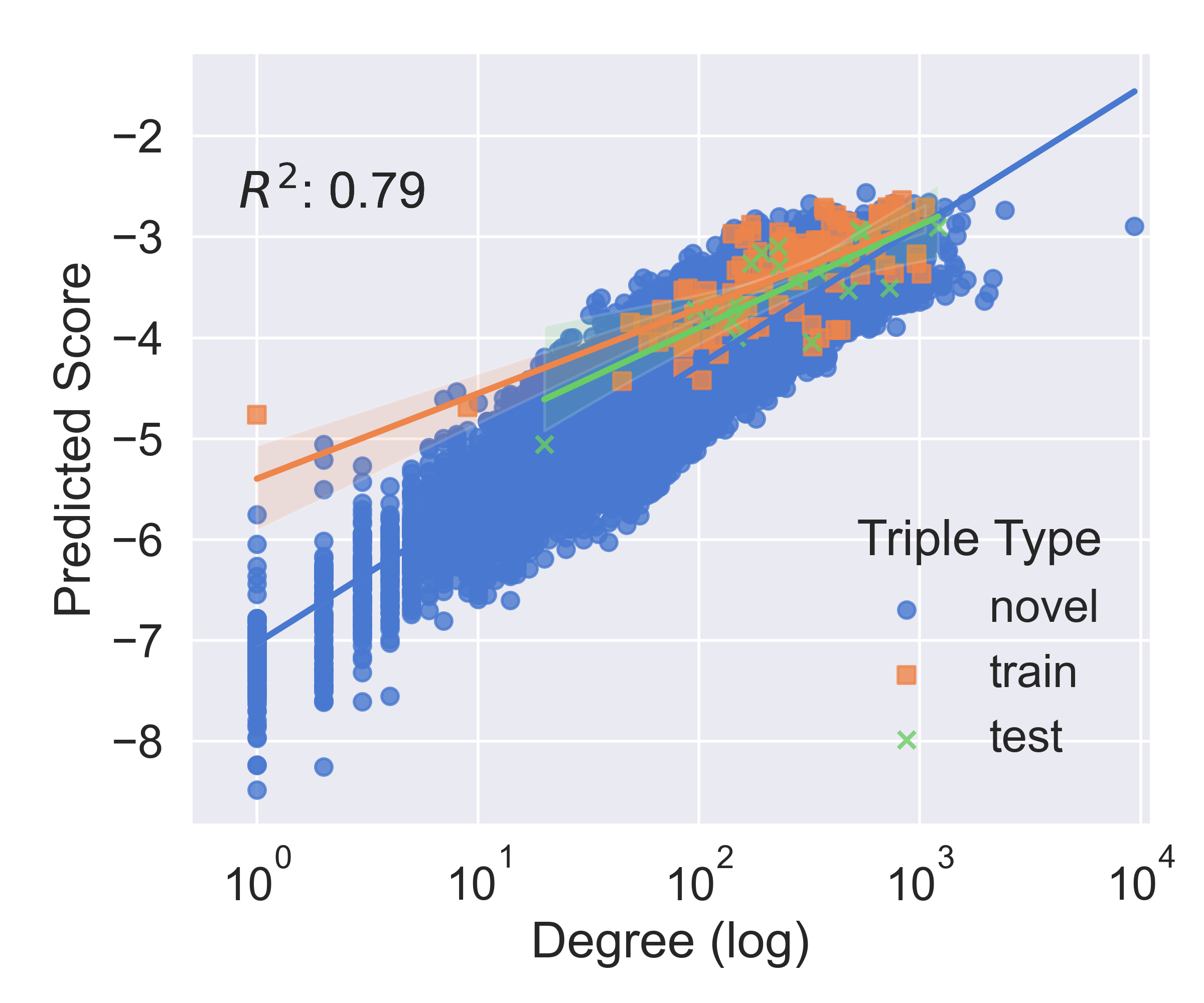}
		\caption{Parkinsons}\label{fig:dr:park}
	\end{subfigure}

	\begin{subfigure}[b]{0.48\textwidth}
		\centering
		\includegraphics[width=0.99\textwidth]{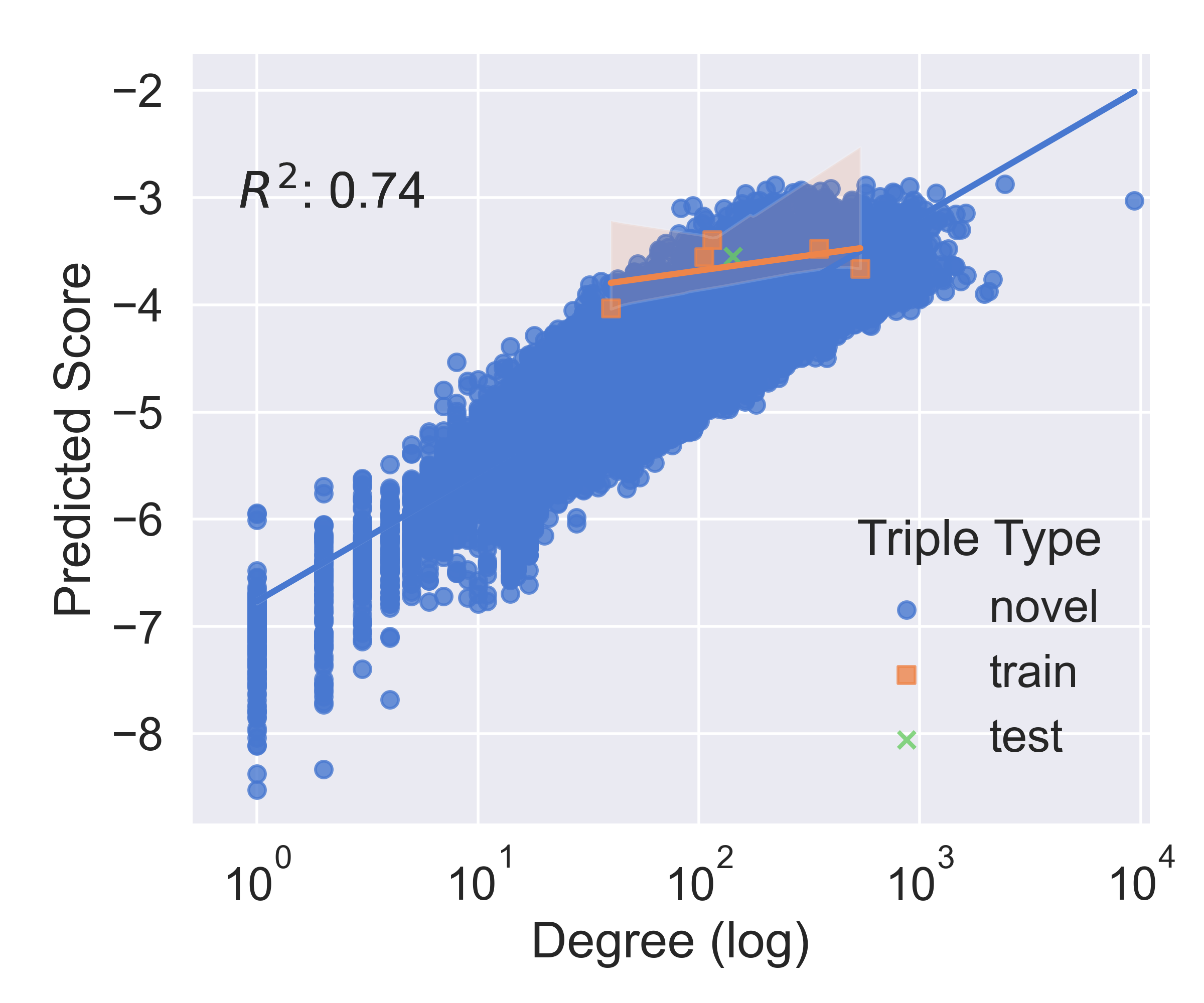}
		\caption{Fuchs Endothelial Dystrophy}\label{fig:dr:fuchs}
	\end{subfigure}
	\begin{subfigure}[b]{0.48\textwidth}
		\centering
		\includegraphics[width=0.99\textwidth]{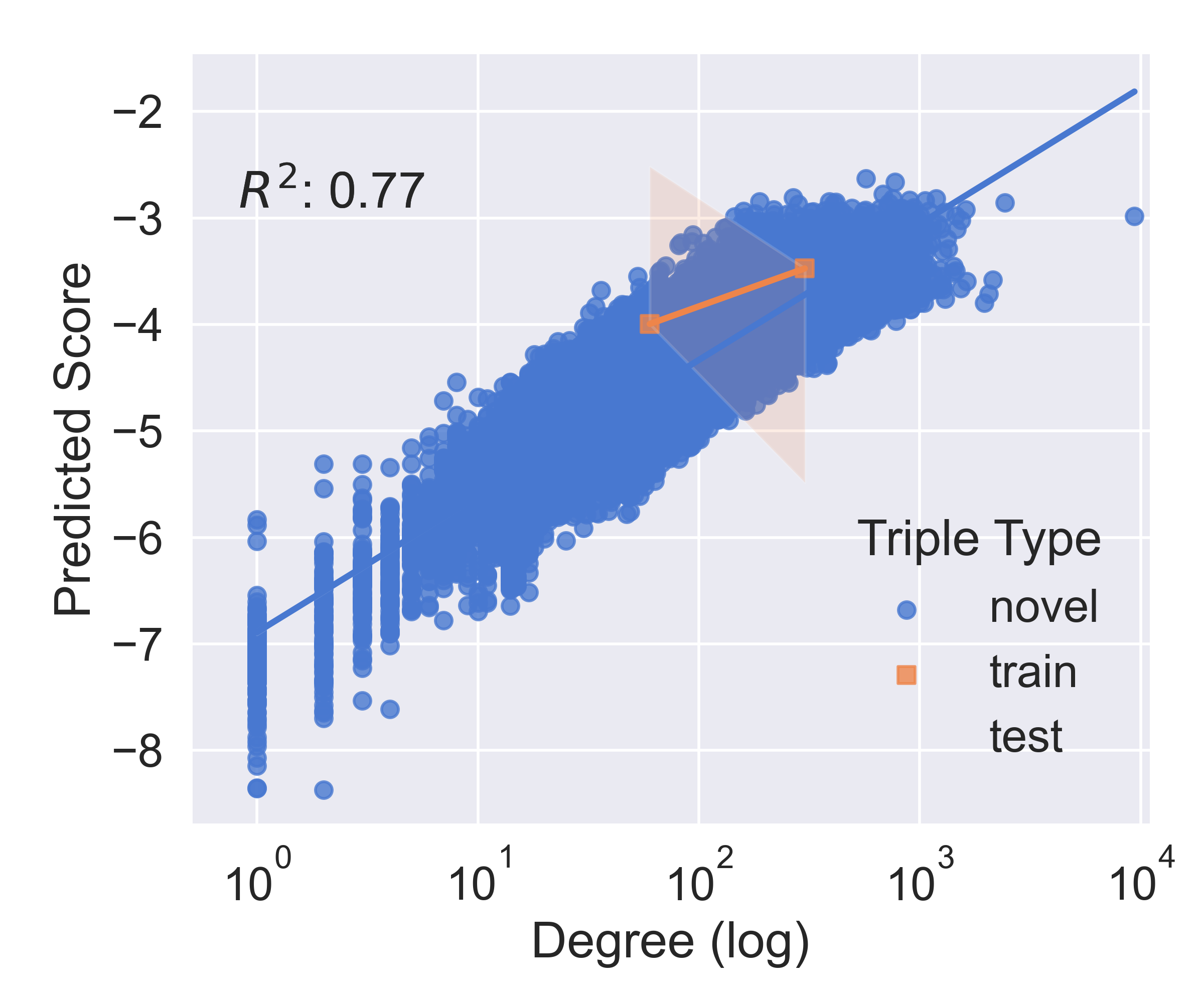}
		\caption{Fallopian Tube Cancer}\label{fig:dr:fal}
	\end{subfigure}
	\caption{Relationship between gene entity degree and the score assigned to it by TransE assessing the likelihood of it associating with the given disease.}\label{fig:degree-regression}
\end{figure*}

\begin{figure}[!ht]
	\centering
	\begin{subfigure}[b]{0.45\textwidth}
		\centering
		\includegraphics[width=0.99\textwidth]{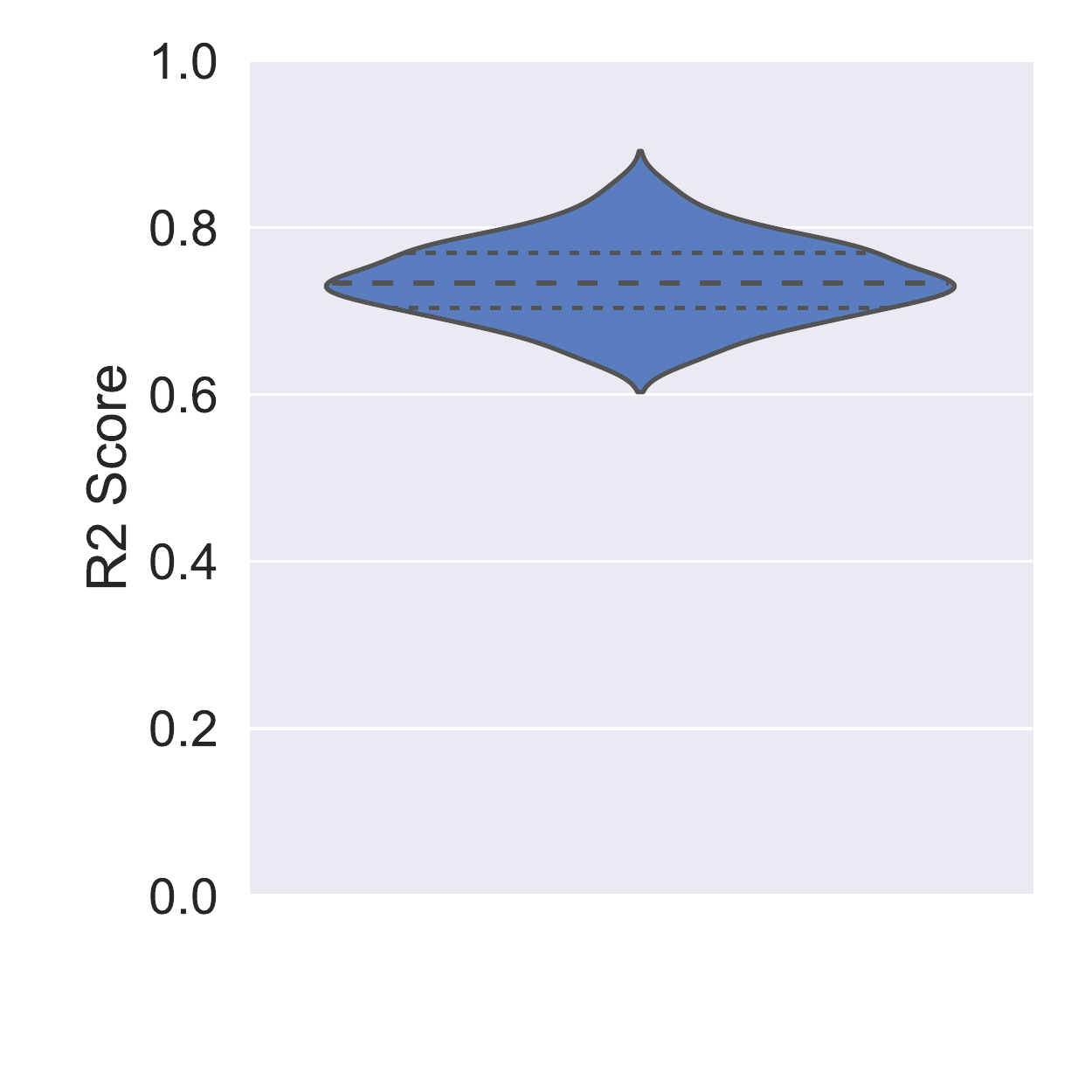}
		\caption{Distribution of \(R^2\)}\label{fig:r2:violin}
	\end{subfigure}
	\begin{subfigure}[b]{0.45\textwidth}
		\centering
		\includegraphics[width=0.99\textwidth]{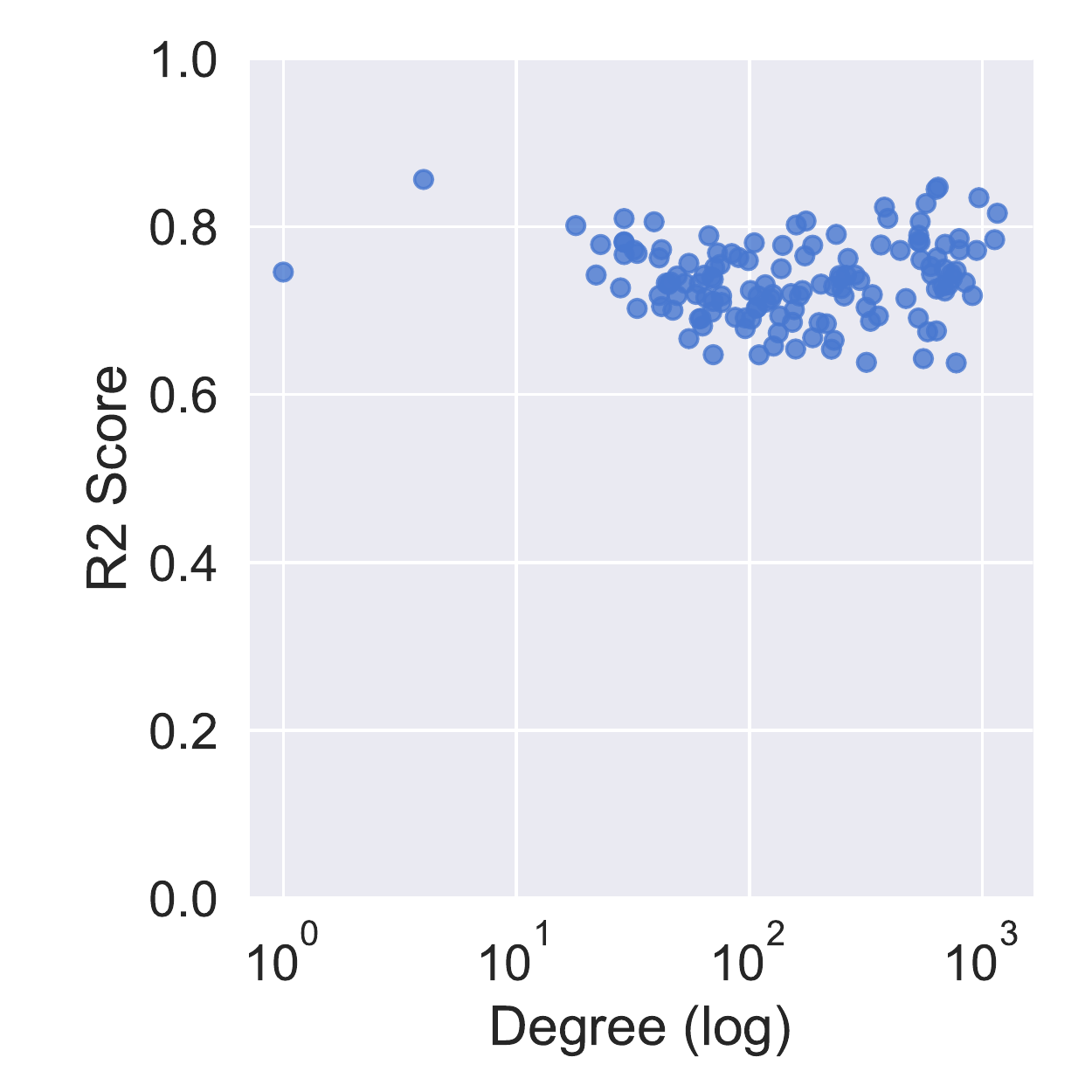}
		\caption{\(R^2\) values vs disease degree}\label{fig:r2:dist}
	\end{subfigure}
	\caption{Correlation between the gene degree and score is consistent across all diseases in Hetionet as shown here. All diseases in Hetionet show high correlation coefficients between gene degree vs score, while the disease degree seems not to influence the results.}
	\label{fig:r2_over_disease}
\end{figure}

In Figure \ref{fig:degree-regression} the relationship between the score assigned to the gene entities by the TransE model and their degree is presented across both low and highly connected diseases. The points are coloured according to whether the particular entity was seen completing the triple in the train or test datasets, or whether it is novel (i.e. a DaG relationship not present in Hetionet between the given disease and that particular gene). The figures show there is a very clear relationship between the degree of the entity and the score assigned to it by the model -- the higher the degree, the more likely the model considers the entity to complete the triple. This observation holds for both well-connected (Figures \ref{fig:dr:mel} and \ref{fig:dr:park}) and sparsely connected diseases (Figures \ref{fig:dr:fuchs} and \ref{fig:dr:fal}). Indeed the pattern is remarkably similar, not only across all the diseases shown here in detail, but also for all 137 unique diseases that exist in Hetionet (Figure \ref{fig:r2_over_disease}), suggesting that the connectivity of the gene has a large impact on whether it is likely to be predicted as associated with a disease, no matter which disease we investigate.

Another observation that can be made from Figure \ref{fig:degree-regression} is that many entities not seen during training have a higher score given by the model than those in the training data. This is especially true for the low connectivity diseases (Figures \ref{fig:dr:fuchs} and \ref{fig:dr:fal}) where the model ranks many genes with high degree over those with an actual connection in the training data. In other words, the model is more confident in a highly-connected gene than a true positive it has seen during training. While it is prudent not to draw big conclusions with so few data points in these diseases, it is nevertheless an important finding that a high-degree node can be ranked higher than a known positive. Indeed, the perceived wisdom would be that the model would over-fit to the training data and one would expect the training points to all be given high scores regardless of their degree, which does not appear to be the case.

Additionally, we wanted to explore how this apparent degree bias impacts the commonly used metrics for ranking entities. We demonstrate this by measuring the Mean Rank (MR) and Mean Reciprocal Rank (MRR), as well as Hits@\textit{k}, with \textit{k}=1 and \textit{k}=10, of the tail entities for all DaG triples present in a holdout testset over 10 different dataset splits, as shown in Table \ref{tab:metrics}. Results are separated into high-degree (genes with degree \(>1000\)) and low-degree (genes with degree \(<200\)). The table shows that the model indeed demonstrates significantly lower performance, often by an order of magnitude, when evaluating on low-degree ground truth data.

\begin{table}[h!]
	\centering

	\begin{tabular}{l  c c c  c c c  }
		\toprule
		\textbf{Degree} & \multicolumn{4}{c}{\textbf{Metric}} \T\B                                                                    \\
		\midrule \midrule
		                & MR \(\downarrow\)                        & MRR \(\uparrow\)  & Hits@1 \(\uparrow\) & Hits@10 \(\uparrow\)\B \\
		\cline{2-5}
		Low             & 4353.7\(\pm\)198.8                       & 0.006\(\pm\)0.001 & 0.001\(\pm\)0.001   & 0.012\(\pm\)0.003\T    \\
		High            & 300.85 \(\pm\)96.6                       & 0.053\(\pm\)0.018 & 0.010\(\pm\)0.017   & 0.118\(\pm\)0.053\B    \\
		\bottomrule
	\end{tabular}
	\vspace{5pt}
	\caption{Ranking metrics over high and low degree DaG edges from the holdout testsets. Arrows indicate if a higher or lower value is best for a given metric.}
	\label{tab:metrics}
\end{table}

\begin{figure}[!th]
	\centering
	\includegraphics[width=0.9\textwidth]{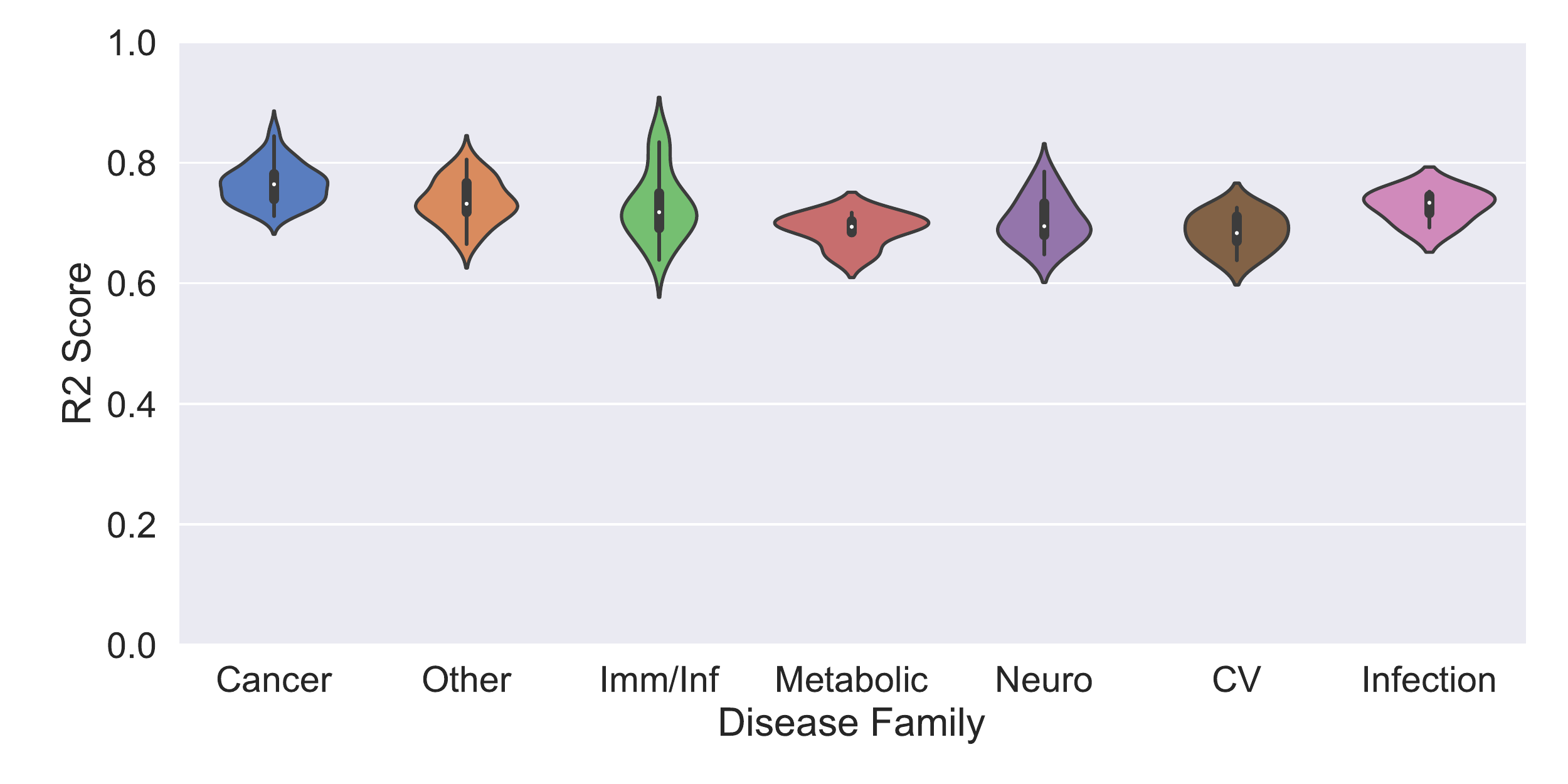}
	\caption[R-Squared value across disease family]{R-Squared value of the relationship between gene degree and score across all 137 diseases in Hetionet categorised into disease families. Disease families are Cancer, Immune/Inflammatory (Imm/Inf), Metabolic, Neurological (Neuro), Cardio-Vascular (CV), Infection and Other.}
	\label{fig:r2-across-disease}
\end{figure}

\textbf{Correlation Across Diseases} Figure \ref{fig:r2-across-disease} shows how the R-Squared value for the correlation between gene degree and predicted score across all 137 diseases in Hetionet, where the diseases have been categorised into families. The figure highlights how the strong correlation is present across the whole range of diseases in the dataset.

\begin{figure*}[!h]
	\centering
	\begin{subfigure}[b]{0.24\textwidth}
		\centering
		\includegraphics[width=0.99\textwidth]{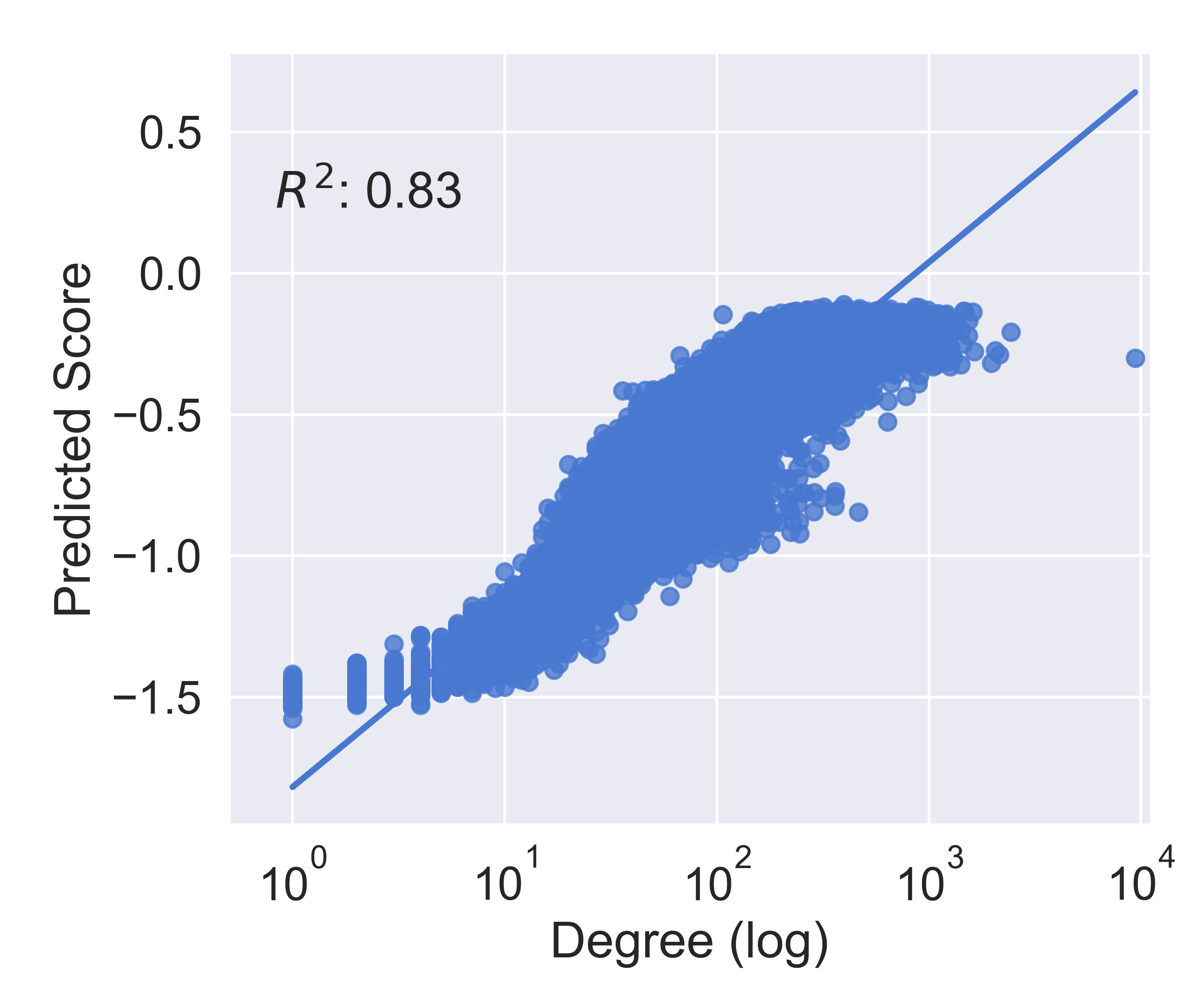}
		\caption{TransH (0.11)}\label{fig:dr:transh}
	\end{subfigure}
	\begin{subfigure}[b]{0.24\textwidth}
		\centering
		\includegraphics[width=0.99\textwidth]{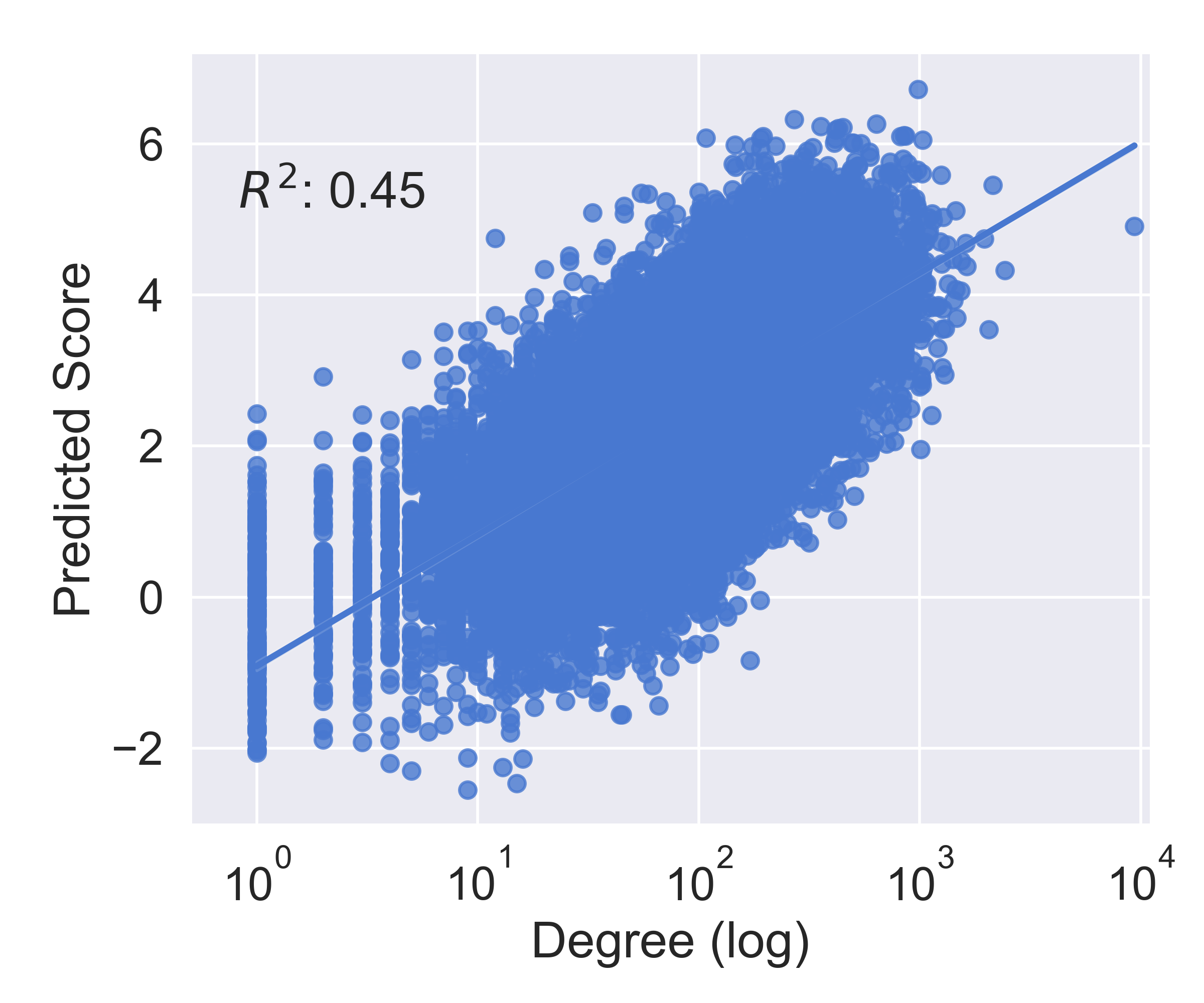}
		\caption{ComplEx (0.07)}\label{fig:dr:complex}
	\end{subfigure}
	\begin{subfigure}[b]{0.24\textwidth}
		\centering
		\includegraphics[width=0.99\textwidth]{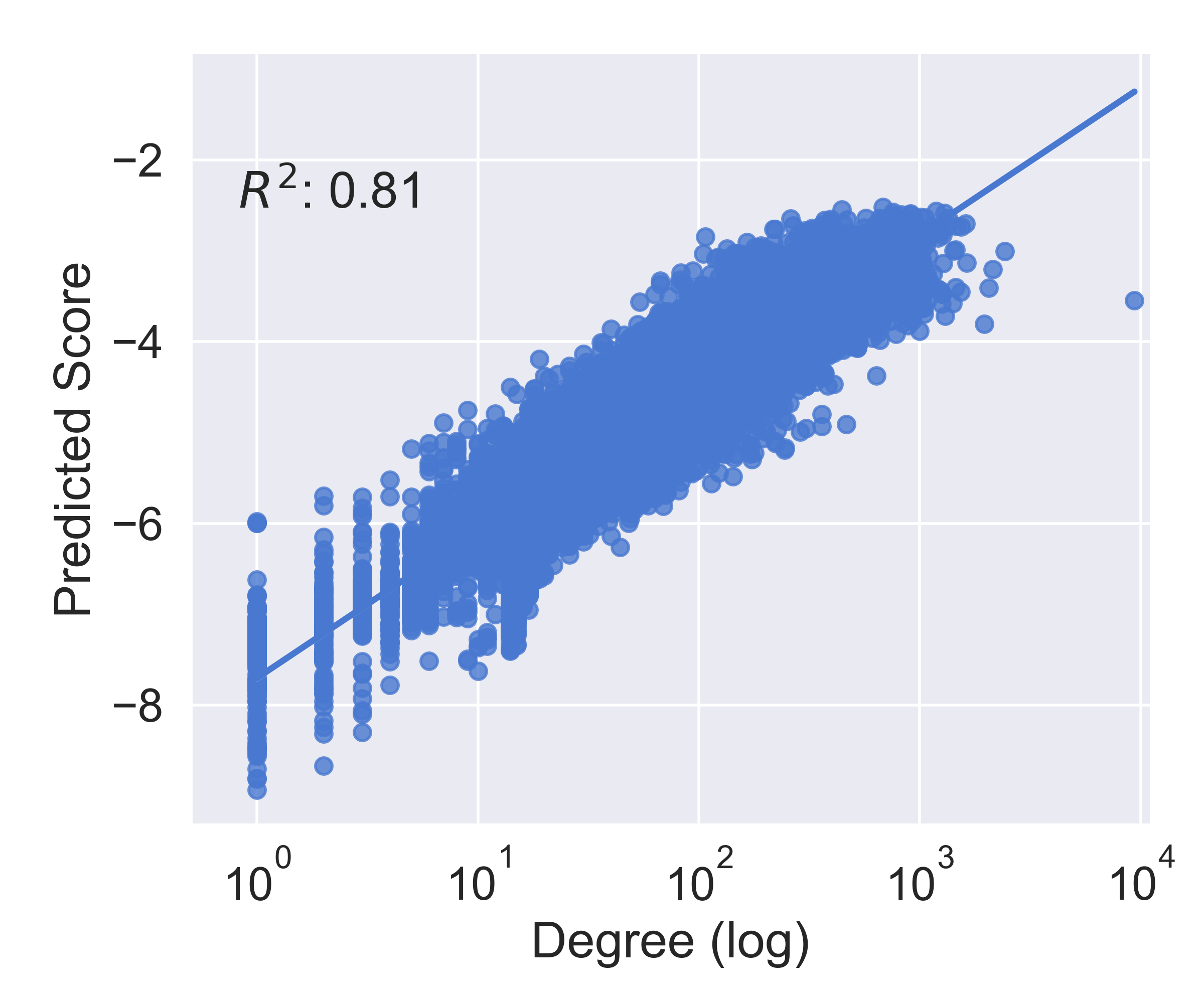}
		\caption{RotatE (0.27)}\label{fig:dr:rotate}
	\end{subfigure}
	\begin{subfigure}[b]{0.24\textwidth}
		\centering
		\includegraphics[width=0.99\textwidth]{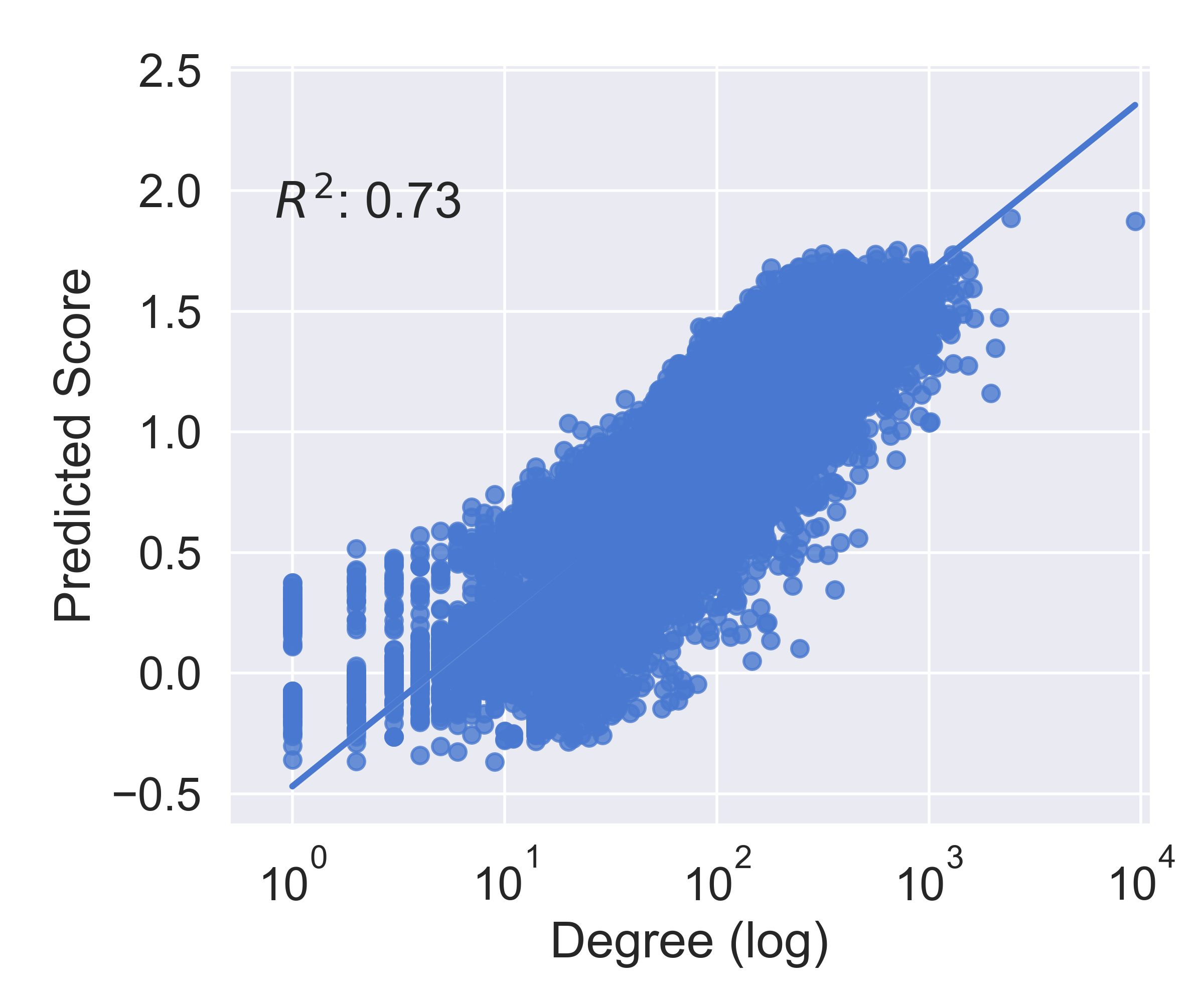}
		\caption{DistMult (0.14)}\label{fig:dr:distmult}
	\end{subfigure}
	\caption{Relationship between gene degree versus predicted score for association with Breast Cancer. Numbers detail the Hits@10 score for that model on a holdout testset. Note that the change scale for the score is due to the different objective function used by the models.}
	\label{fig:degree-regression-models}
\end{figure*}

\textbf{Different Models.} Figure \ref{fig:degree-regression-models} highlights that this relationship between degree and score is not just a TransE specific issue but is prevalent across many knowledge graph embedding models used in the literature (including TransH~\cite{wang2014knowledge}, ComplEx~\cite{trouillon2016complex}, RotatE~\cite{sun2019rotate} \& DistMult~\cite{yang2015embedding} which were trained using optimised hyper-parameters for Hetionet as detailed previously~\cite{bonner2021understanding}). The figure also displays the Hits@10 score on a hold-out testset comprising 10\% of the triples for all models. There appears to be little direct relationship between the Hits@10 score and the R-squared value, although the worse performing model ComplEx does have the weakest correlation between degree and the score assigned to the gene entities.

\begin{figure*}[!h]
	\centering
	\begin{subfigure}[b]{0.32\textwidth}
		\centering
		\includegraphics[width=0.99\textwidth]{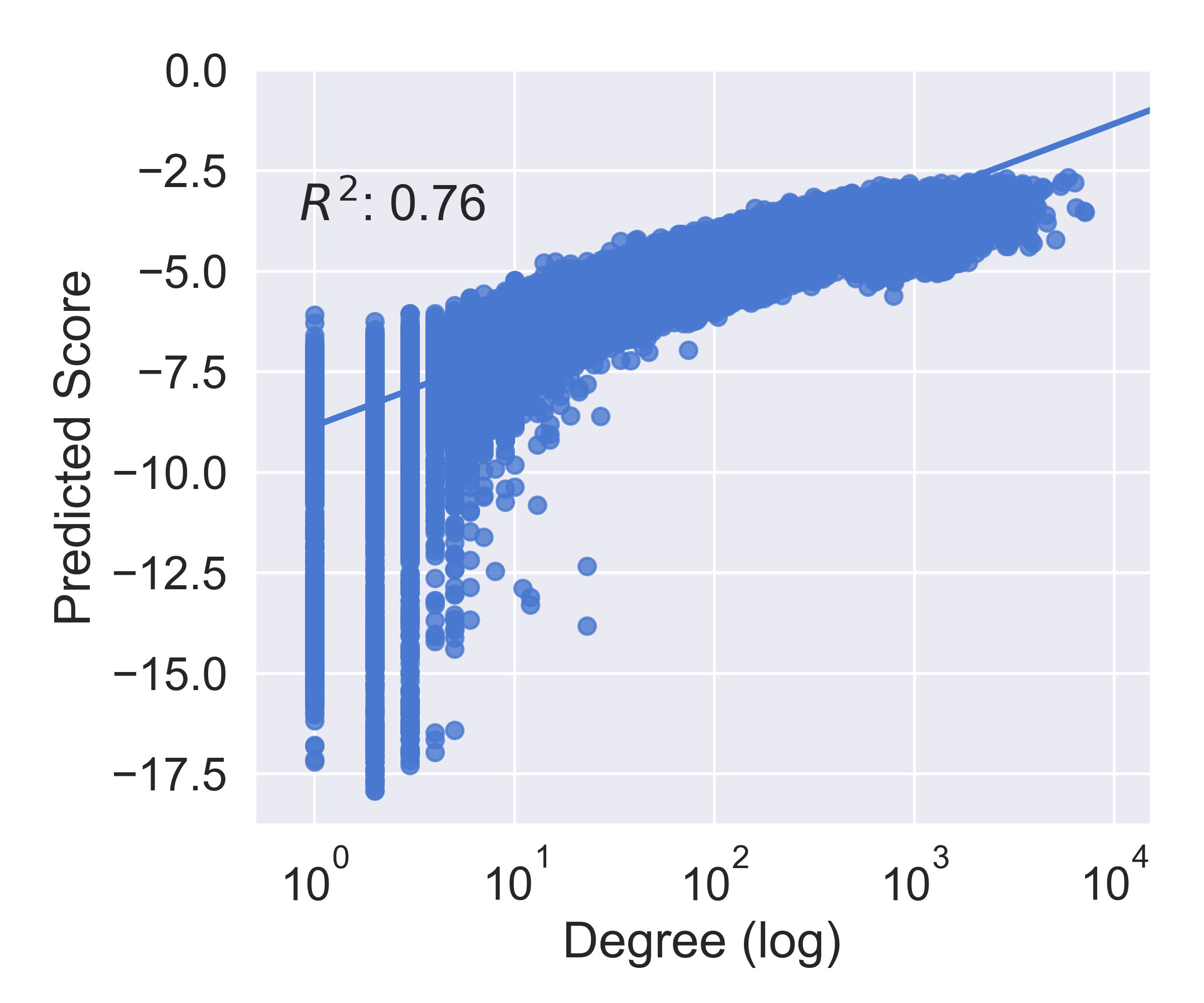}
		\caption{DRKG}\label{fig:other-datasets:drkg}
	\end{subfigure}
	\begin{subfigure}[b]{0.32\textwidth}
		\centering
		\includegraphics[width=0.99\textwidth]{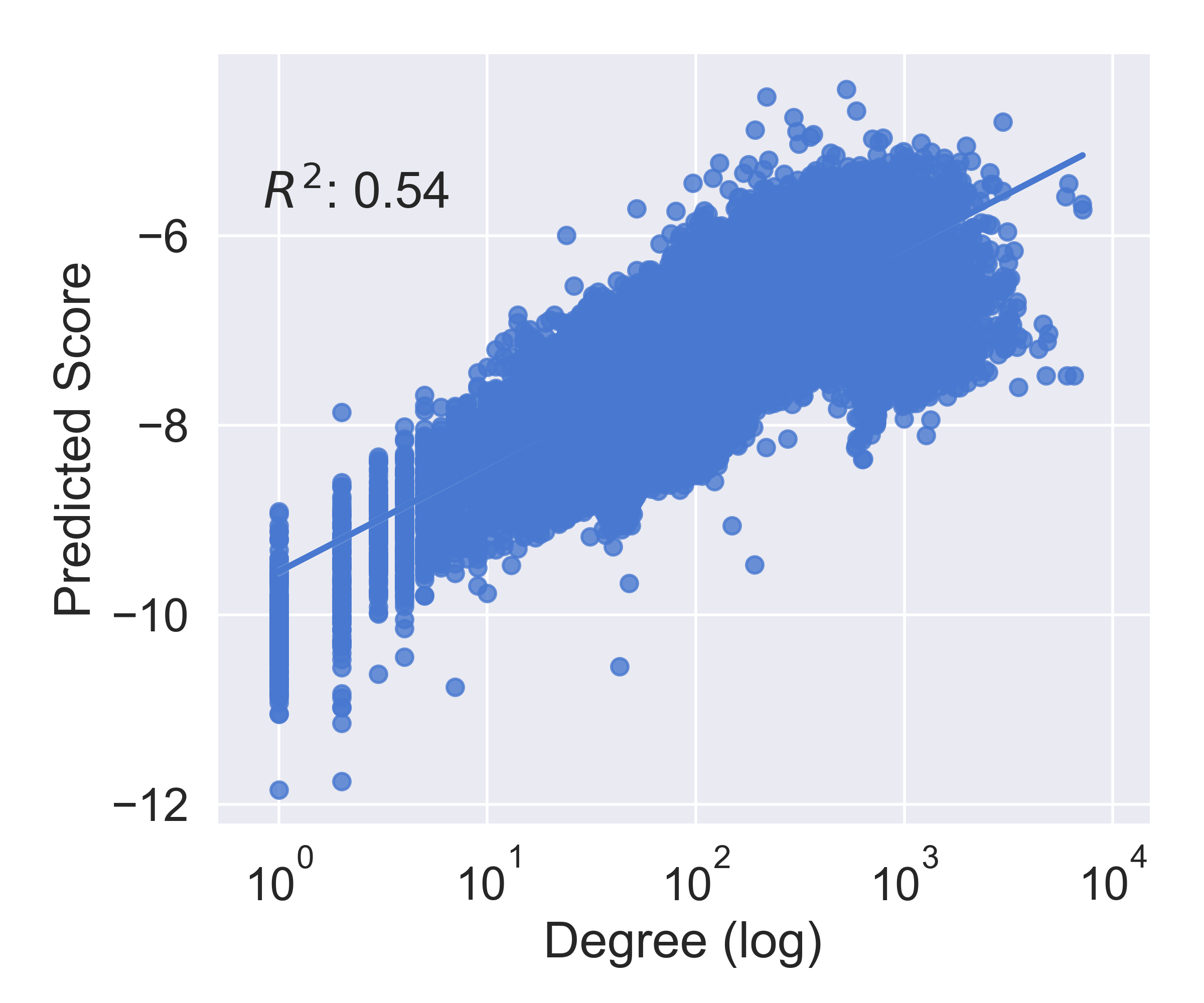}
		\caption{OpenBioLink}\label{fig:other-datasets:obl}
	\end{subfigure}
	\caption{Relationship between the predicted score for gene association with breast cancer and degree across other biomedical knowledge graphs.}
	\label{fig:other-datasets}
\end{figure*}

\textbf{Different Datasets.} Further, Figure \ref{fig:other-datasets} shows this issue not to be a dataset specific one as the same pattern can be found on two other biomedical knowledge graphs: the Drug Repurposing Knowledge Graph (DRKG)~\cite{drkg2020} and OpenBioLink~\cite{breit2020openbiolink}. Here the TransE model is used to predict the genes most likely to be associated with breast cancer. Figure \ref{fig:other-datasets} shows that a correlation between predicted score and degree is indeed present in both datasets, although OpenBioLink has a weaker correlation. Although we only present results for breast cancer for these datasets, we observed similar patterns across all of our chosen diseases.

\begin{figure*}[!ht]
	\centering
	\begin{subfigure}[b]{0.3\textwidth}
		\centering
		\includegraphics[width=0.99\textwidth]{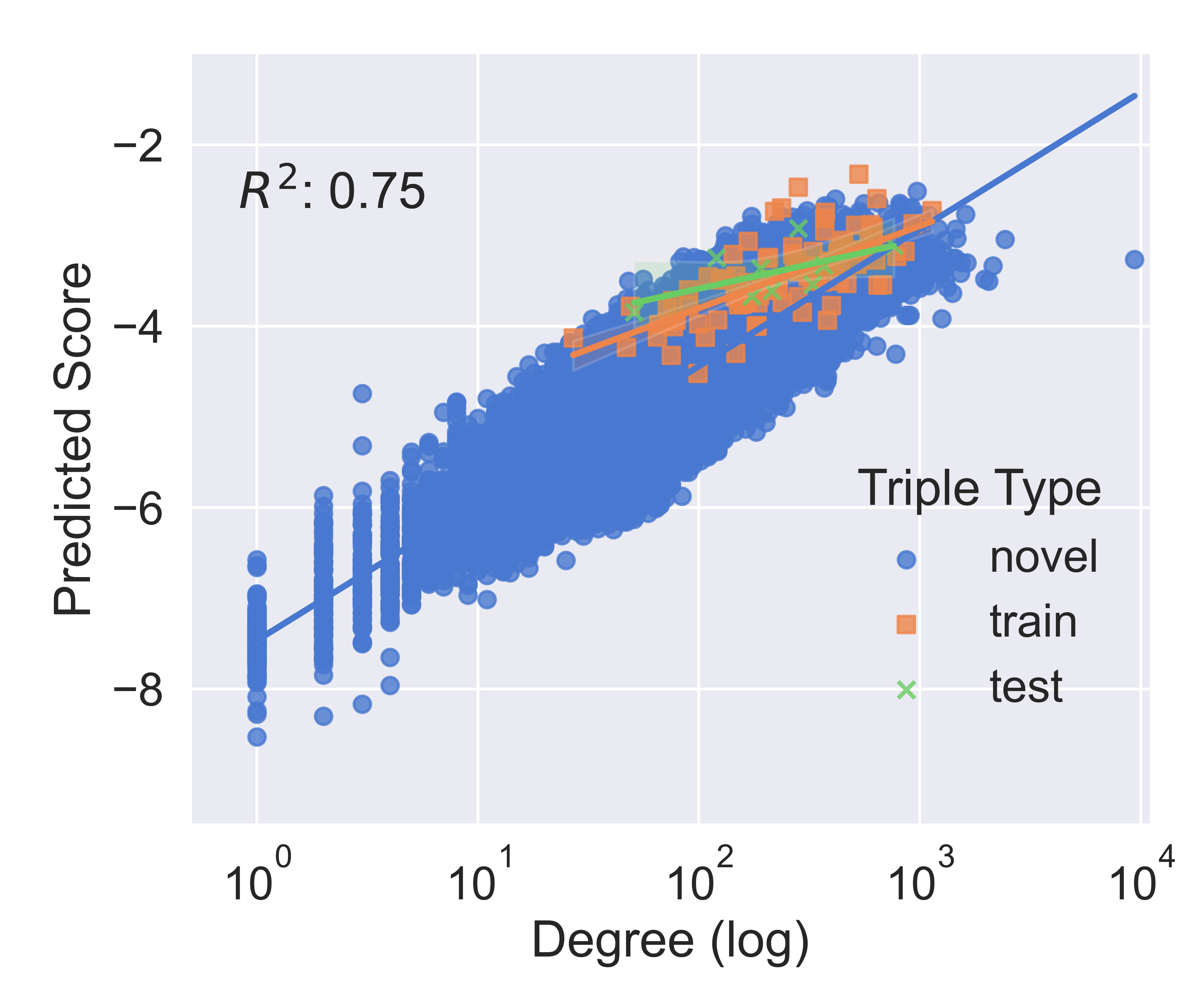}
		\caption{DTI - Sunitinib}\label{fig:other-tasks:dti1}
	\end{subfigure}
	\begin{subfigure}[b]{0.3\textwidth}
		\centering
		\includegraphics[width=0.99\textwidth]{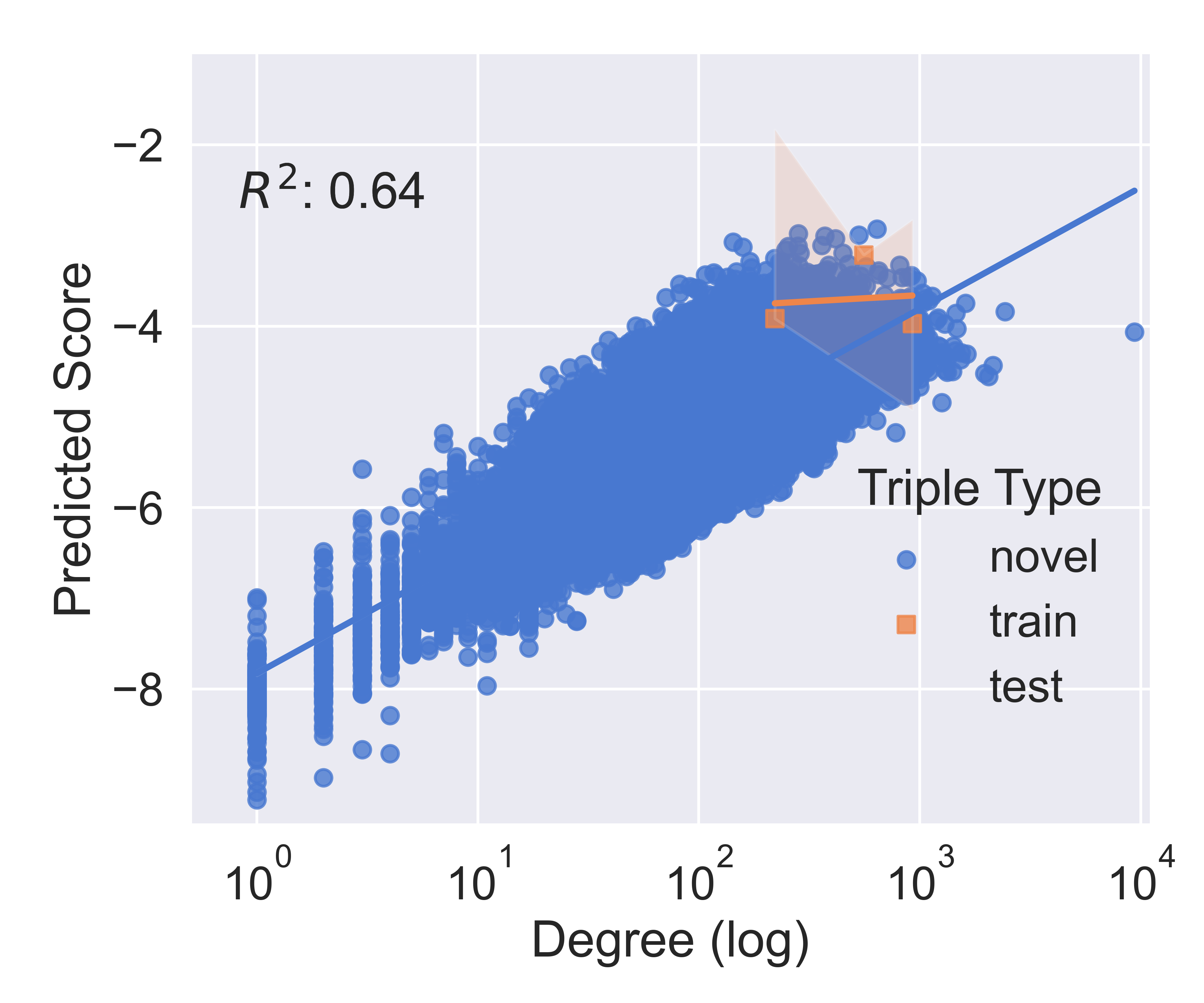}
		\caption{DTI - Auranofin}\label{fig:other-tasks:dti2}
	\end{subfigure}
	\begin{subfigure}[b]{0.3\textwidth}
		\centering
		\includegraphics[width=0.99\textwidth]{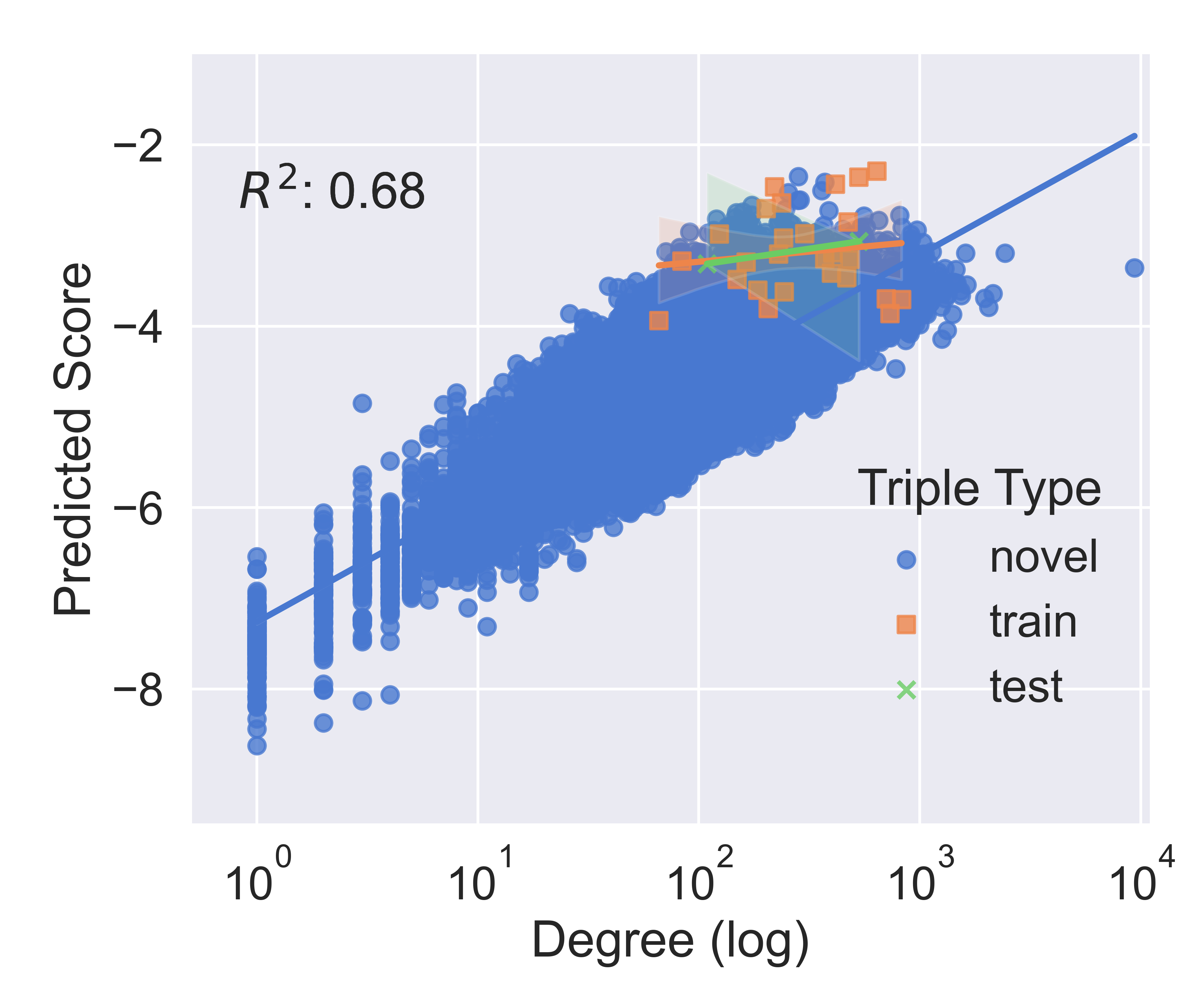}
		\caption{DTI - Doxorubicin}\label{fig:other-tasks:dti3}
	\end{subfigure}

	\begin{subfigure}[b]{0.3\textwidth}
		\centering
		\includegraphics[width=0.99\textwidth]{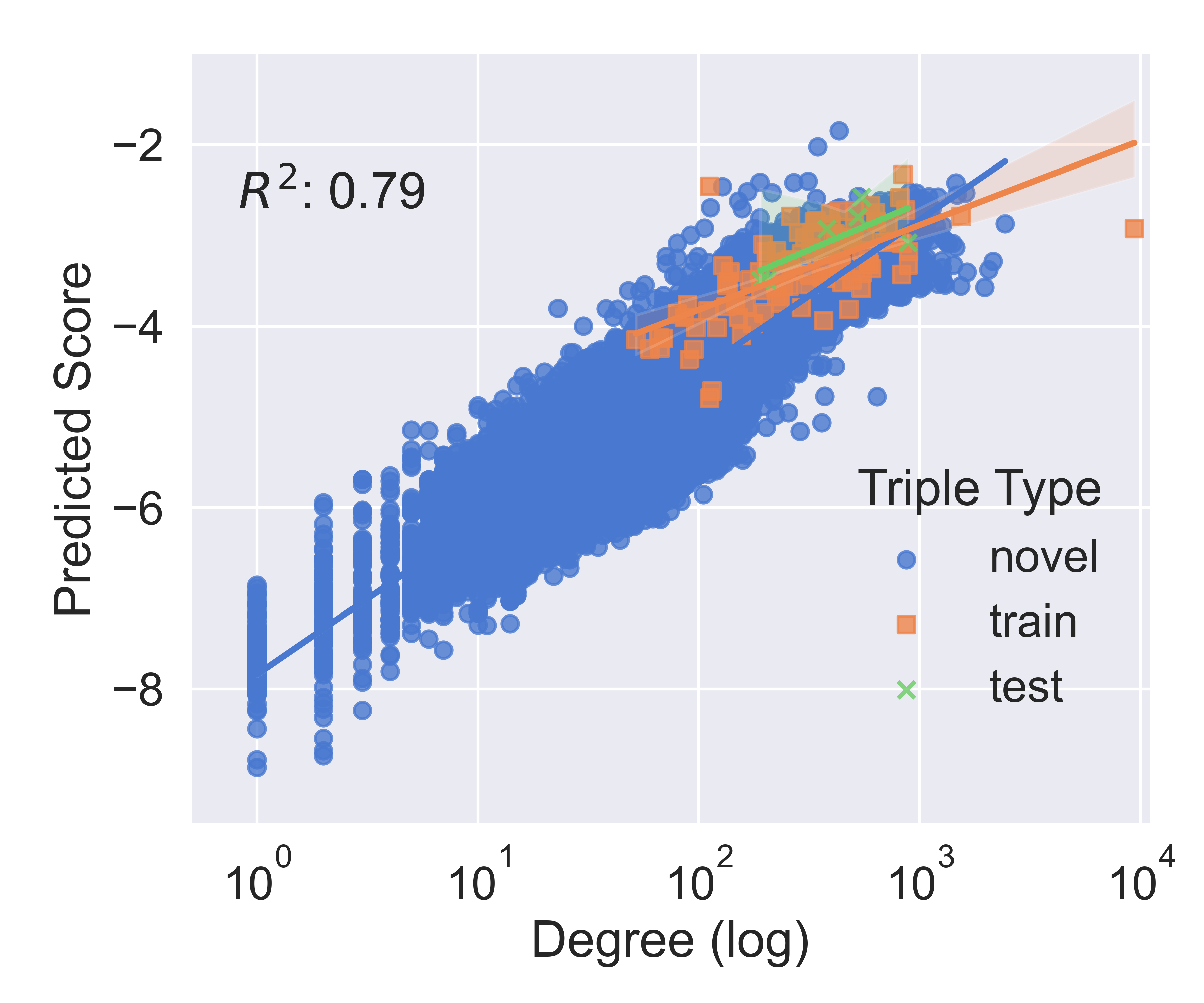}
		\caption{PPI - MAPK1}\label{fig:other-tasks:ppi1}
	\end{subfigure}
	\begin{subfigure}[b]{0.3\textwidth}
		\centering
		\includegraphics[width=0.99\textwidth]{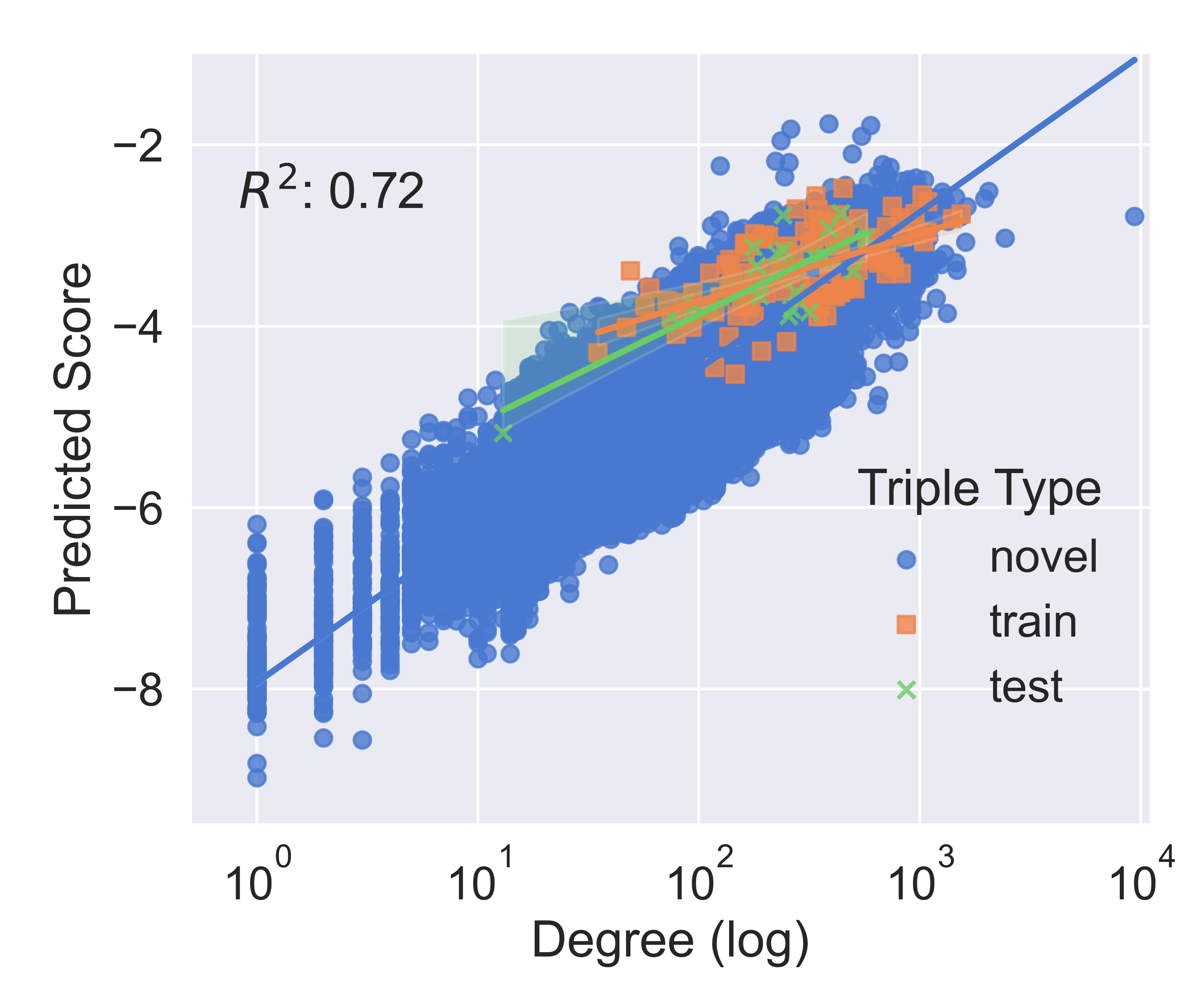}
		\caption{PPI - PCNA}\label{fig:other-tasks:ppi2}
	\end{subfigure}
	\begin{subfigure}[b]{0.3\textwidth}
		\centering
		\includegraphics[width=0.99\textwidth]{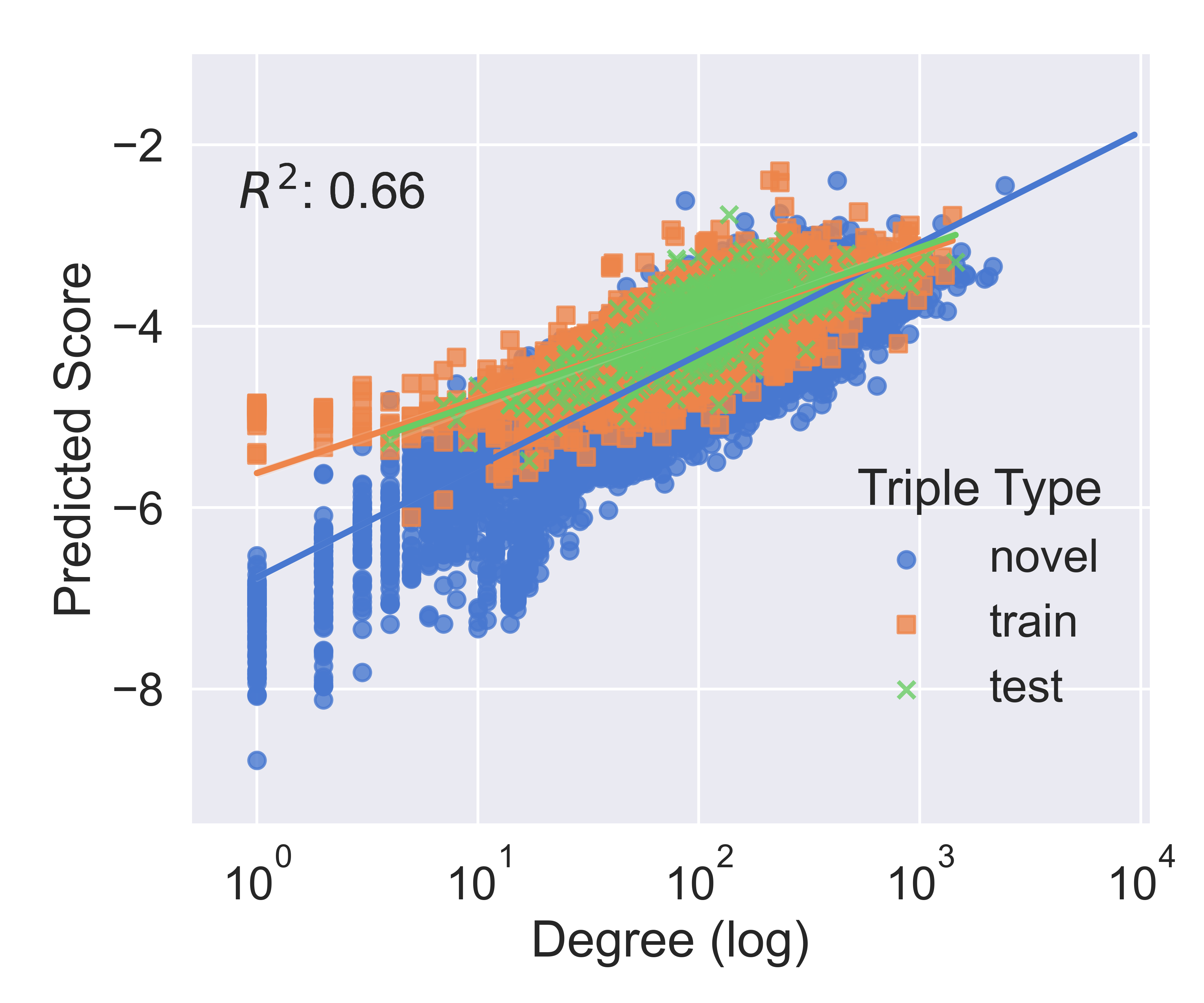}
		\caption{PPI - UBC}\label{fig:other-tasks:ppi3}
	\end{subfigure}

	\begin{subfigure}[b]{0.3\textwidth}
		\centering
		\includegraphics[width=0.99\textwidth]{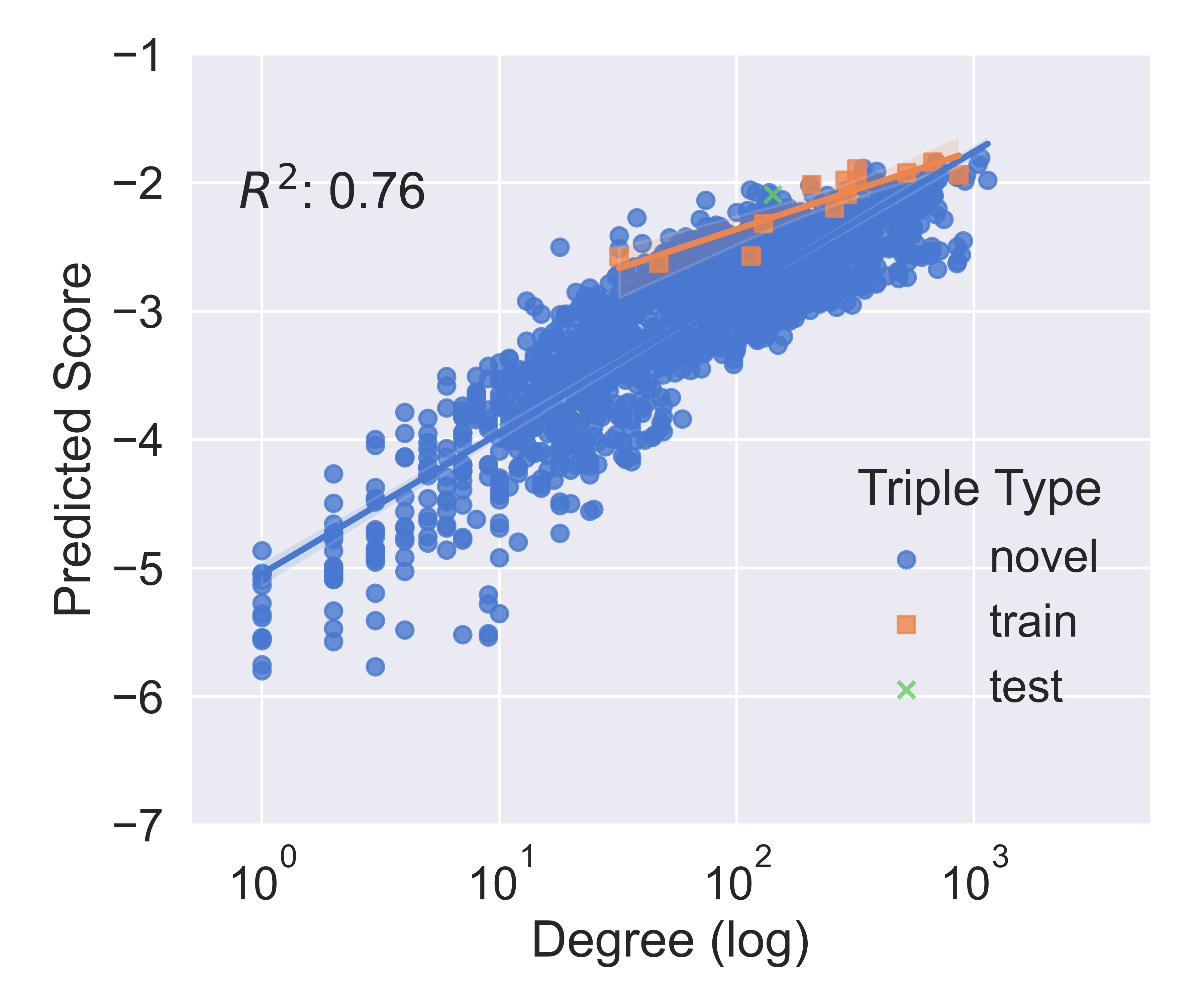}
		\caption{DR - Rheumatoid arthritis}\label{fig:other-tasks:ctd1}
	\end{subfigure}
	\begin{subfigure}[b]{0.3\textwidth}
		\centering
		\includegraphics[width=0.99\textwidth]{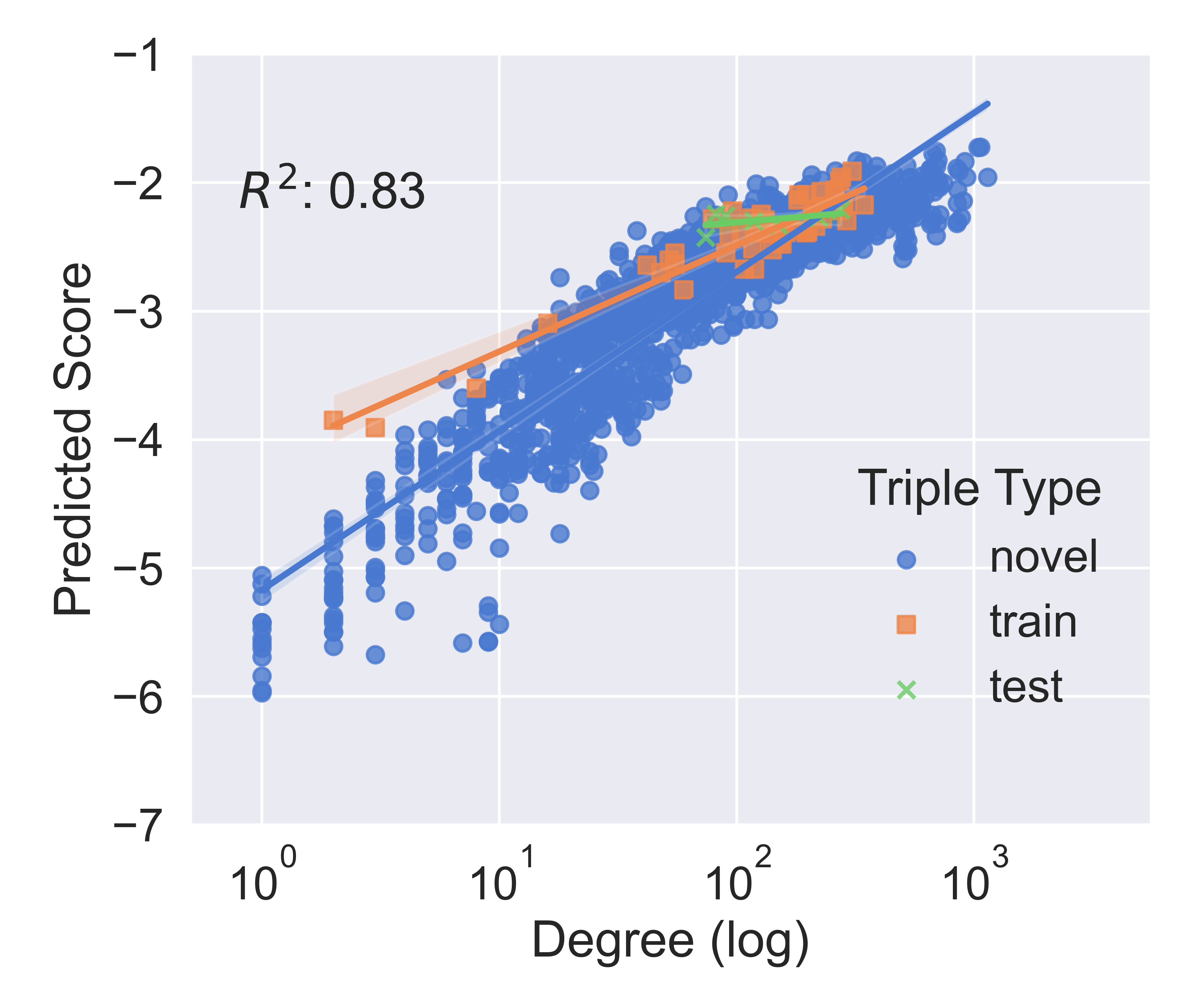}
		\caption{DR - Hypertension}\label{fig:other-tasks:ctd2}
	\end{subfigure}
	\begin{subfigure}[b]{0.3\textwidth}
		\centering
		\includegraphics[width=0.99\textwidth]{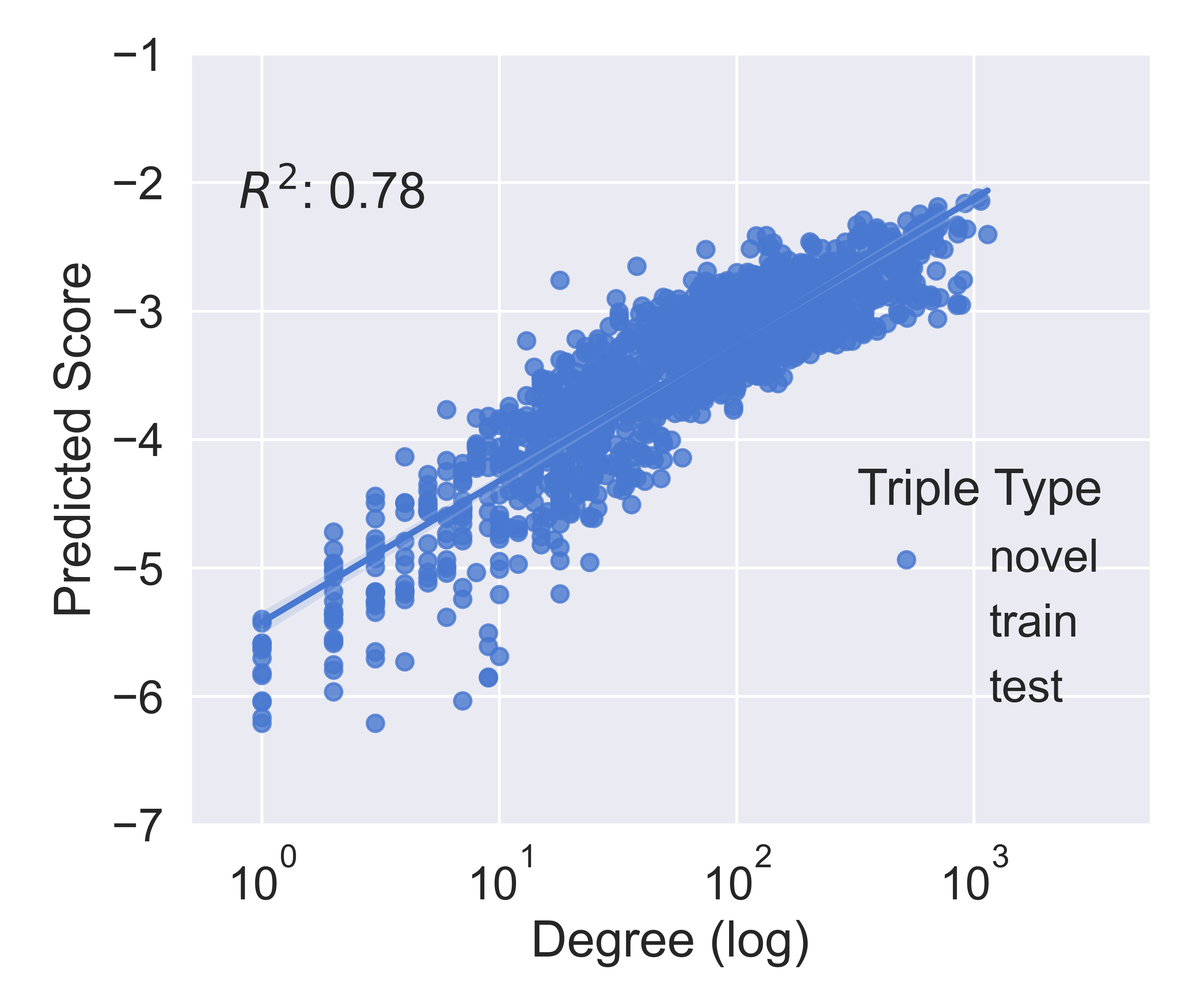}
		\caption{DR - FED}\label{fig:other-tasks:ctd3}
	\end{subfigure}
	\caption{Relationship between the prediced score and degree for other drug discovery tasks;  in a-c we query drug-target interactions for three compounds, d-f protein-protein interaction partners for three genes  are queried, and in g-i we query for compounds for treatment of three diseases.}
	\label{fig:other-tasks}
\end{figure*}

\textbf{Different Tasks.} We show that the issue is not specific to the task of gene-disease association prediction, but is replicated across many drug discovery tasks. Figure \ref{fig:other-tasks} highlights the relationship between score and entity degree for three other tasks, namely: Drug Target Interaction (DTI), Protein-Protein Interaction (PPI) and Drug Repurposing (DR). For the DTI task (Figures \ref{fig:other-tasks:dti1}-\ref{fig:other-tasks:dti3}), we use the \(\mathit{Compound} \xrightarrow[]{\mathit{binds}} \mathit{Gene}\) (CbG) edge types to rank genes by how likely they are to bind to the drugs Sunitinib, Auranofin and Doxorubicin. These compounds are selected in order to showcase different aspects of the dataset. Sunitinib and Doxorubicin are both chemotheraphy agents against cancer; the former is a receptor tyrosine kinase while the latter is an anthracycline topoisomerase inhibitor. Both drugs are well studied and thus well connected in the graph, Sunitinib has the most CbG connections, while Doxorubicin is extremely well connected both in terms of overall degree, and specifically connections to genes. On the opposing end of the connectivity scale Auranofin has fewer overall connections in the graph, and very few CbG connections thus provides a good example when there is little information to train on.

In the PPI task (Figures \ref{fig:other-tasks:ppi1}-\ref{fig:other-tasks:ppi3}), the \(\mathit{Gene} \xrightarrow[]{\mathit{interacts}} \mathit{Gene}\) (GiG) edge types are used and genes are ranked based on likelihood to interact with the query genes MAPK1, PCNA and UBC. These are all fairly well studied genes, despite their apparent differences in terms of degree in Hetionet. An overwhelming majority of the connectivity of UBC is to other genes, specifically of the type GiG we consider here. PCNA on the other hand has again majority of its connections to other genes, however these are of a different relationship type. Lastly, MAPK1 has a much more balanced connectivity profile, in terms of diversity to other entities in the graph. Thus it provides a good test case to see if connectivity to other node types make any tangible contribution to the predictive performance in GiG relationships.

Finally, in the DR task (Figures \ref{fig:other-tasks:ctd1}-\ref{fig:other-tasks:ctd3}), the \(\mathit{Compound} \xrightarrow[]{\mathit{treats}} \mathit{Disease}\) edges are utilised to rank compounds by how likely the model considerers them to treat the diseases Rheumatoid arthritis, Fuchs endothelial dystrophy and Hypertension. This choice of diseases is motivated by two well-connected diseases of different aetiologies, both with various compounds for treatment options currently used in the clinic. The third disease here is a weakly connected one, without any compound connectivity and surgery as a primary means of treatment. Thus it provides a negative control for the predictive performance for the model. Last but not least, all three diseases are chosen to be distinct from various forms of cancer, as these disease dominate the overall connectivity, as well as to compounds in particular. Many cancer drugs are used for various indications and indeed a model could simply do well by extrapolating a well-connected compound to a well-connected form of cancer.

Across all tasks and example cases, we observe a strong correlation between score and degree highlighting that this is not a task-specific issue and is prevalent across many drug discovery problems.

\subsection{Relations with Disease}
\label{ssec:rel-disease}

We now consider whether the number of connections a gene has to disease entities within the training graph can affect ranking.

\begin{figure*}[!ht]
	\centering
	\begin{subfigure}[b]{0.4\textwidth}
		\centering
		\includegraphics[width=0.99\textwidth]{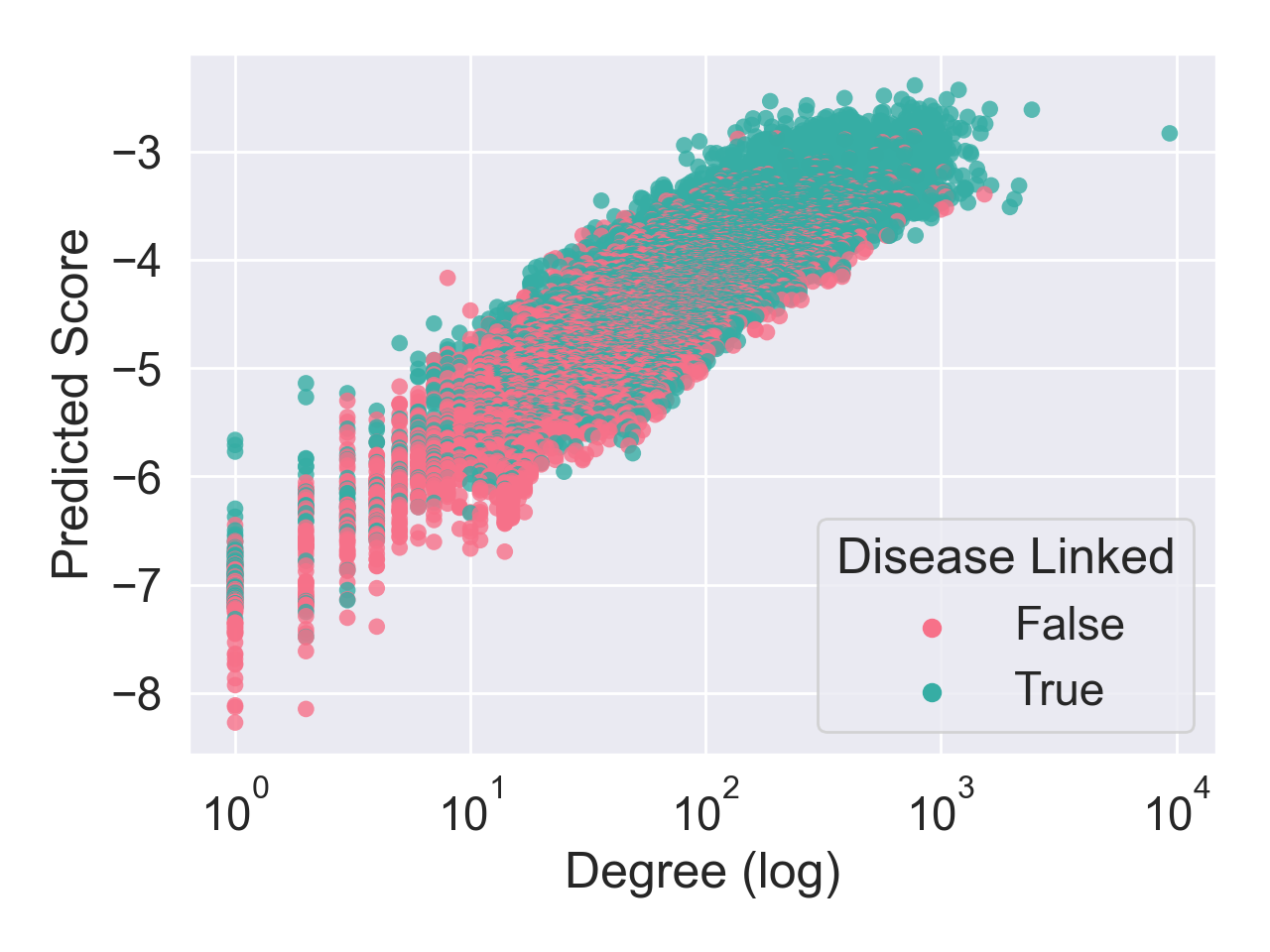}
		\caption{Disease Linked}\label{fig:disease:deg}
	\end{subfigure}
	\begin{subfigure}[b]{0.4\textwidth}
		\centering
		\includegraphics[width=0.74\textwidth]{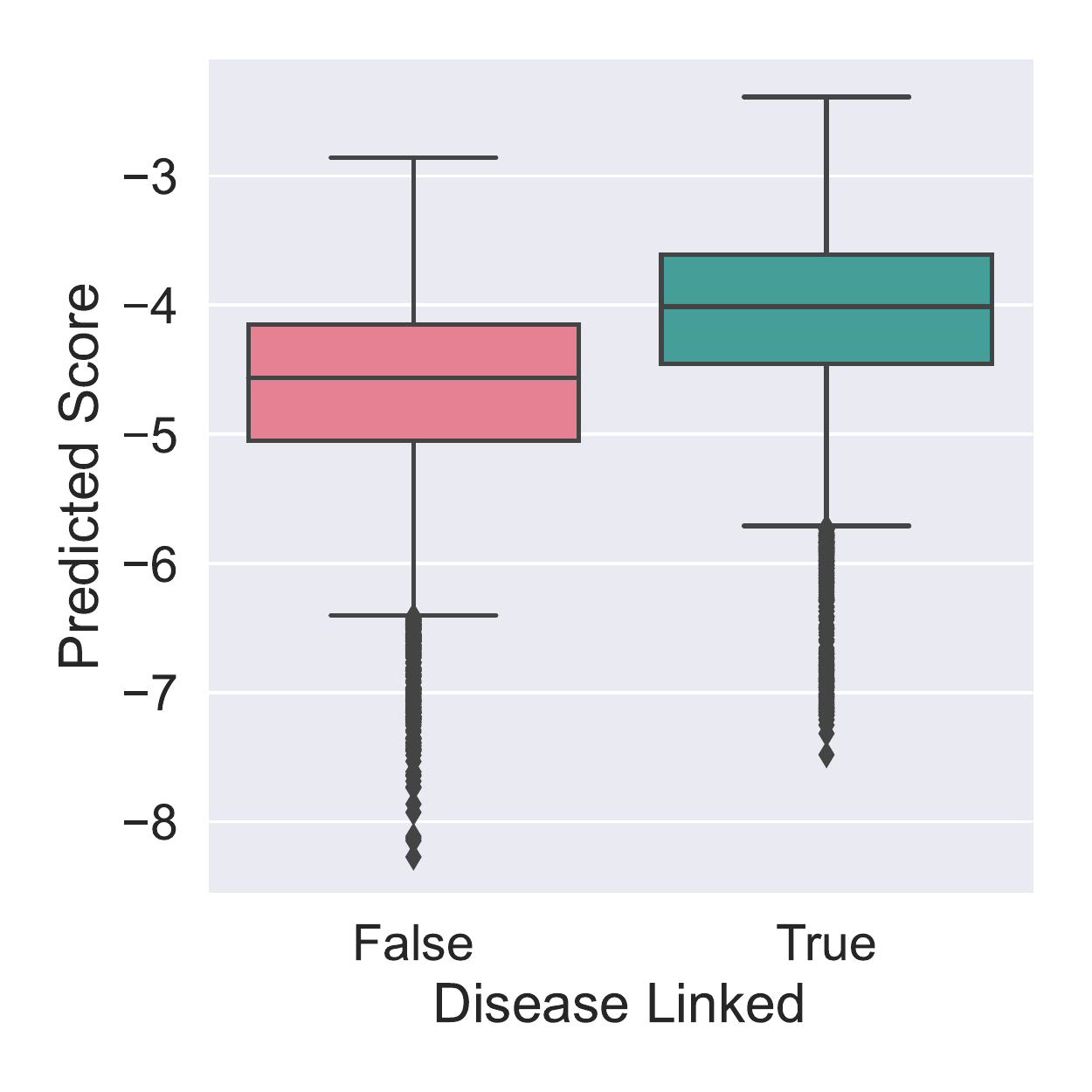}
		\caption{Disease Linked (Box)}\label{fig:disease:deg-box}
	\end{subfigure}

	\begin{subfigure}[b]{0.4\textwidth}
		\centering
		\includegraphics[width=0.99\textwidth]{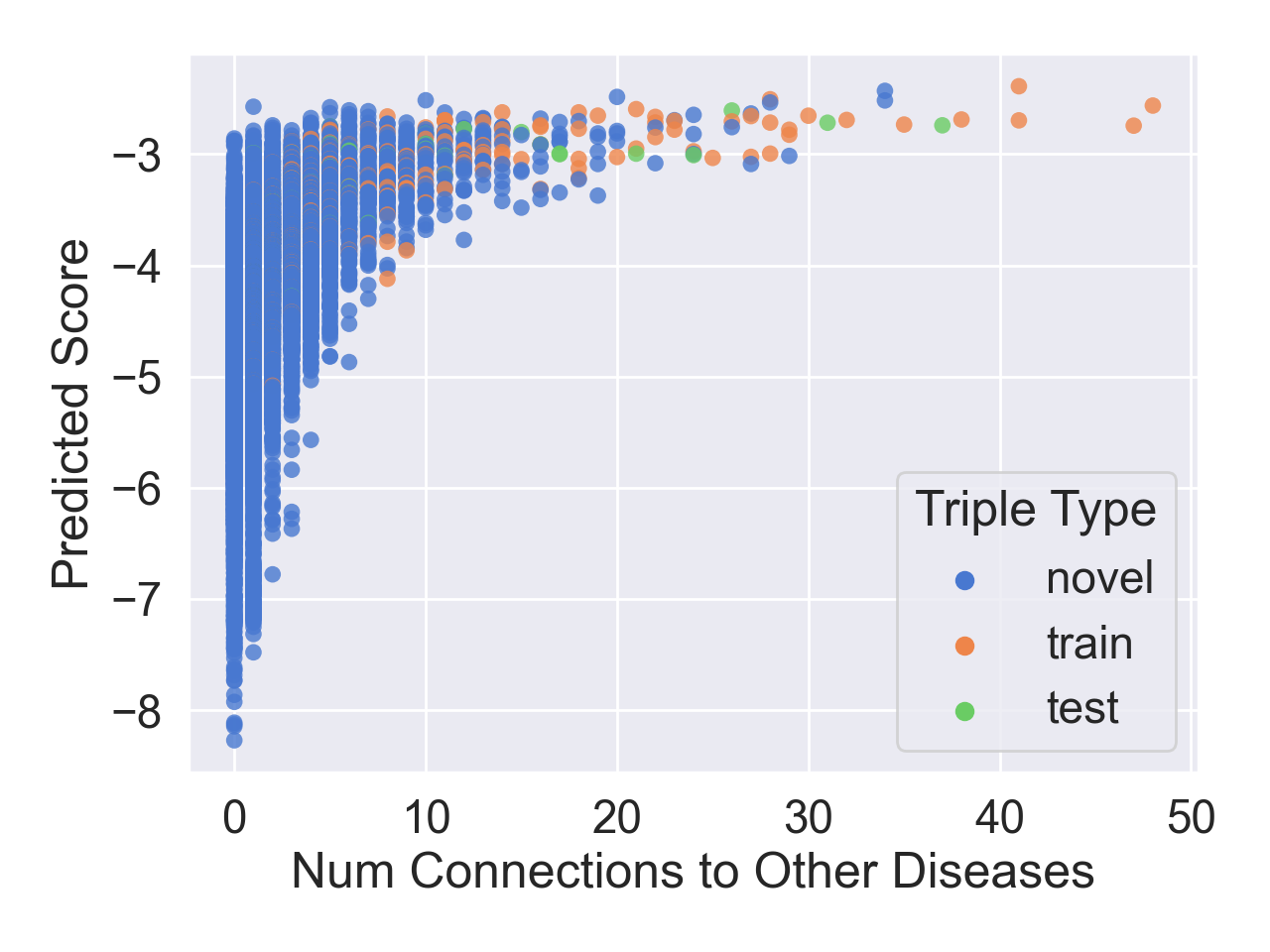}
		\caption{Number of disease connections}\label{fig:disease:num}
	\end{subfigure}
	\begin{subfigure}[b]{0.4\textwidth}
		\centering
		\includegraphics[width=0.99\textwidth]{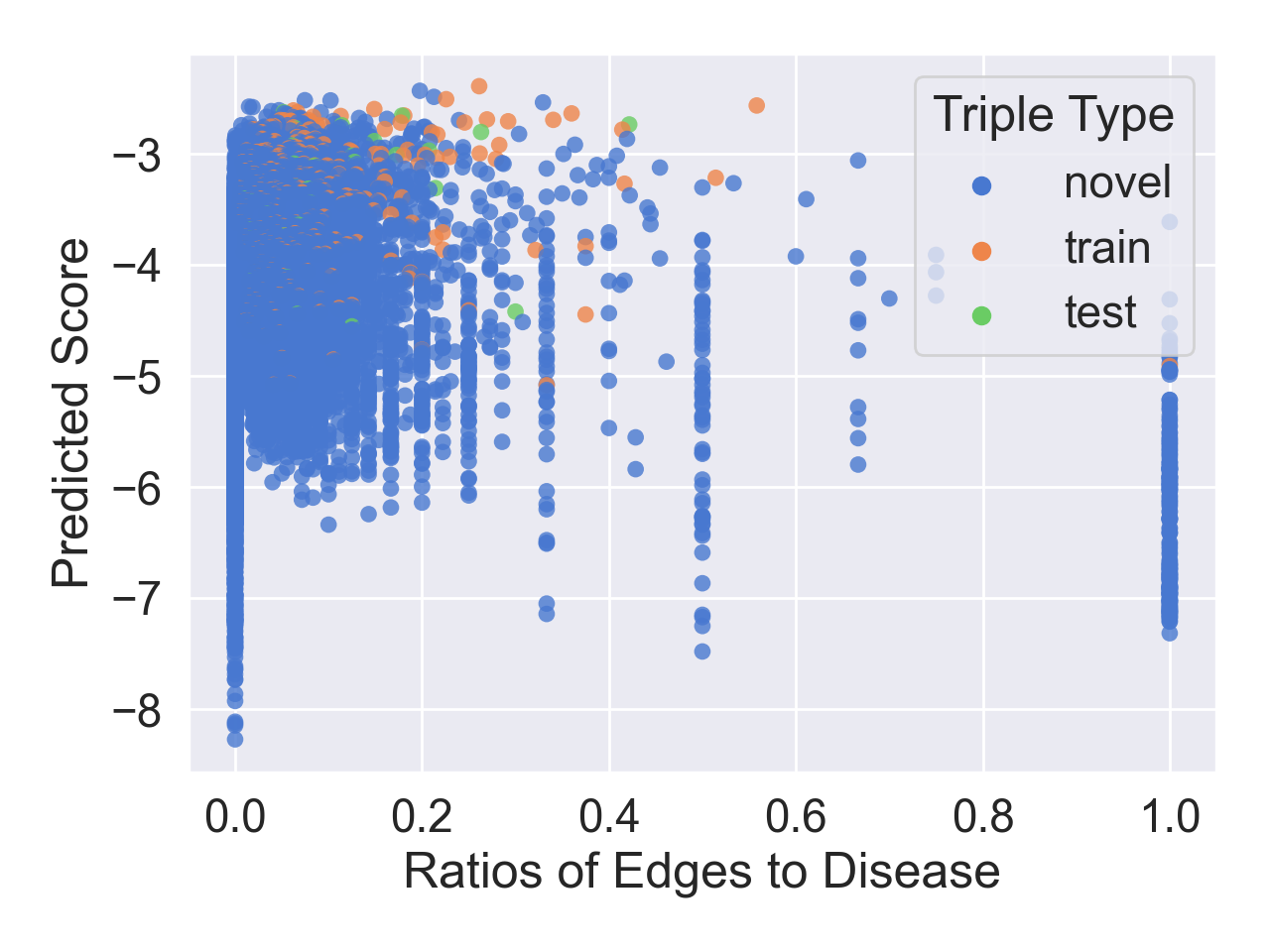}
		\caption{Ratio of disease edges}\label{fig:disease:ratio}
	\end{subfigure}
	\caption{The relationship between the prediced score for Melanoma and various measures of connectivity between genes and the disease entities.}
	\label{fig:disease-connections}
\end{figure*}

Figure \ref{fig:disease-connections} shows this analysis where melanoma is the query disease in question. Intuitively, one would expect a gene that has multiple connections to diseases to have a higher likelihood of being associated with another disease. This follows from the fact that in order to be associated with any disease in the first place, the gene would need to be characterized and often these associations are experimentally validated. Therefore, it is more likely that a well-characterized gene is associated with another disease, than a completely unknown gene is to be linked to a disease for the first time.

As expected, we found that the genes that have links to disease entities tend to have a higher predicted score (and thus higher rank) than those that have no disease links, as shown in Figure \ref{fig:disease:deg}. As mentioned above, gene-disease annotations are not uniformly distributed over genes. In other words, some genes have many disease annotations while others have few or none. Considering this, we wanted to check whether or not we can see a difference between high-ranking and low-ranking genes in terms of total number of connections to other diseases (Figure \ref{fig:disease:num}).

Overall, we observed there is a connection between the number of edges to other diseases and the ranking, such that genes with a higher number of connections to other diseases, tend to have higher scores. However, not all high ranking genes had a large number of edges, which brings us to the question of whether it is the proportion of the edges of a gene to disease nodes that play a significant role. This did not turn out to be the case, as shown in Figure \ref{fig:disease:ratio} there is little relationship between the proportion of edges a gene has to disease entities and the score assigned to it. However, ultimately it is hard to disentangle the impact of disease connections versus the total number, as highly connected genes are also more likely to be connected to diseases.

\subsection{Trivial Relations}\label{ssec:trivial-relations}

We now consider whether the presence of other relation types between a given gene and disease pair in the training graph affects score.

\begin{figure*}[!ht]
	\centering
	\begin{subfigure}[b]{0.48\textwidth}
		\centering
		\includegraphics[width=0.99\textwidth]{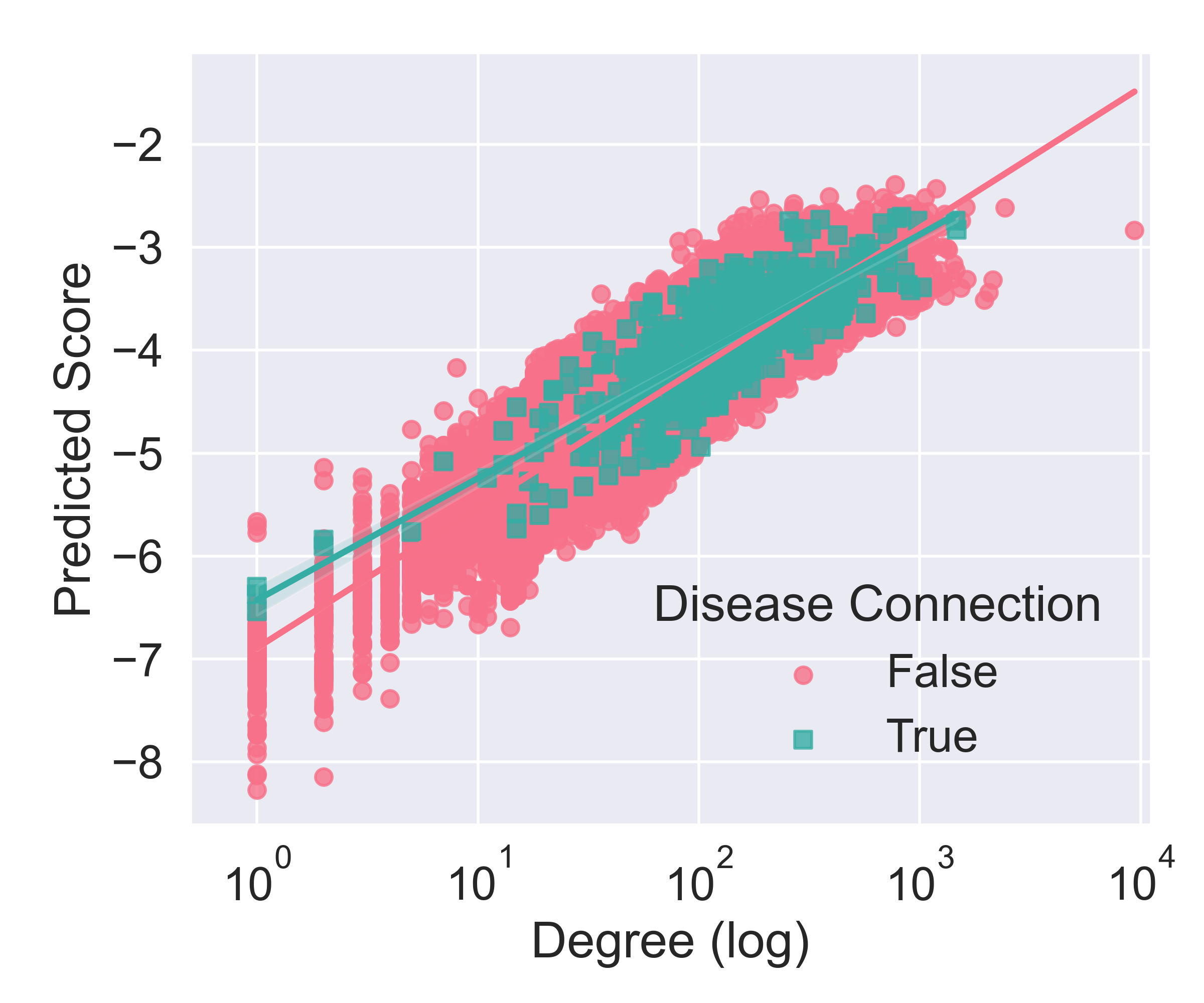}
		\caption{Melanoma}\label{fig:disease:mel}
	\end{subfigure}
	\begin{subfigure}[b]{0.48\textwidth}
		\centering
		\includegraphics[width=0.99\textwidth]{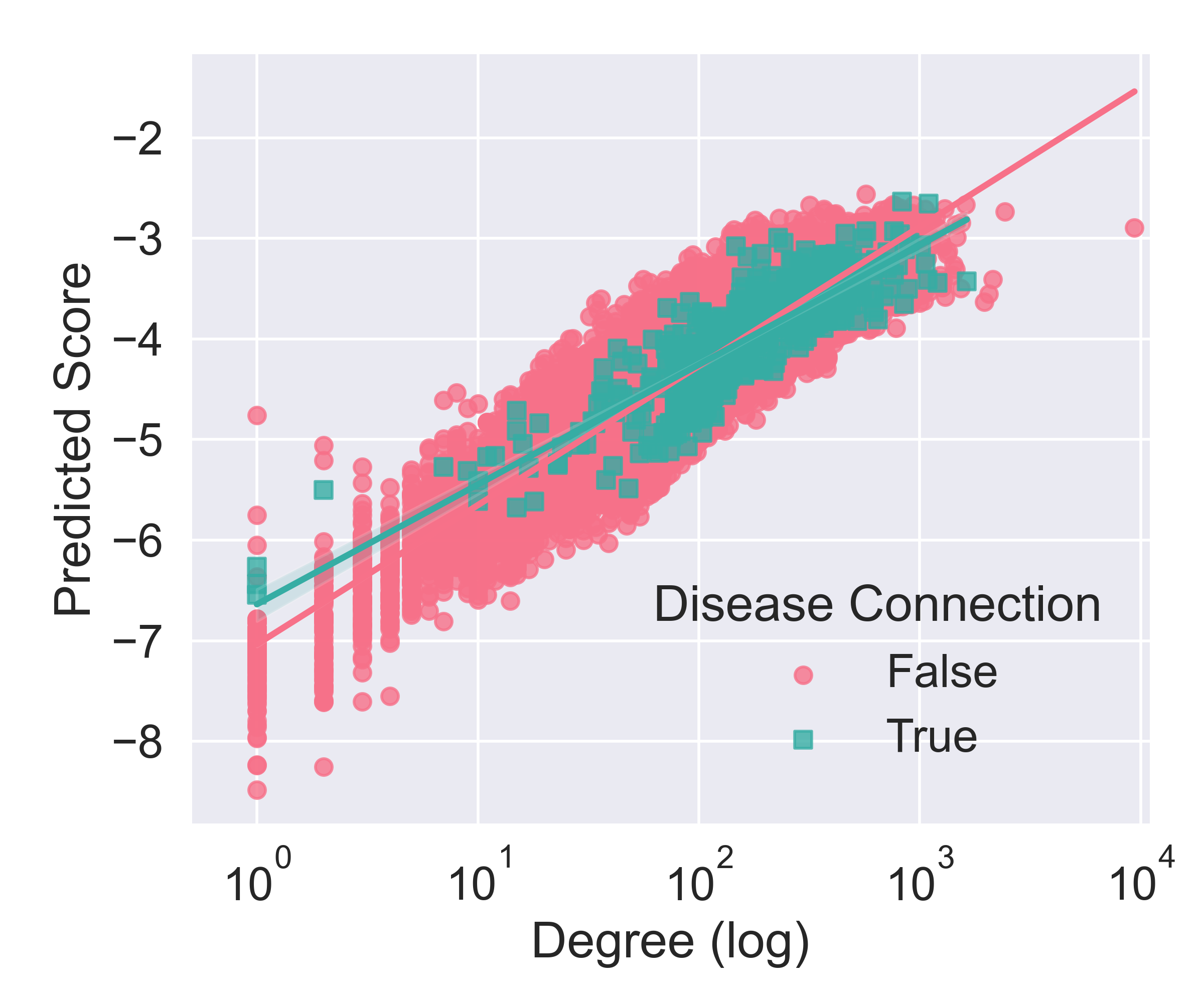}
		\caption{Parkinsons}\label{fig:disease:park}
	\end{subfigure}
	\caption{Relationship between predicted gene-disease association score and entity degree. Points are coloured to indicate whether the gene has an edge of a different type to the disease of interest.}
	\label{fig:disease-other-edge}
\end{figure*}

This is a subtle but important aspect, as multiple relationship types between entities in biological KGs like Hetionet typically allow for information leakage in the form of trivialized predictions. The presence of trivial relations, such as inverse ones, is known to result in over-optimistic models, which although appear as though they are demonstrating good predictive performance, in reality have exploited trivial patterns in the data~\cite{toutanova2015observed, dettmers2018convolutional}. In Hetionet besides the DaG relation type we have used so far, there are also two other relationship types that are allowed between disease and gene entities, specifically \(\mathit{Disease} \xrightarrow[]{\textit{upregulates}} \mathit{Gene}\) and \(\mathit{Disease} \xrightarrow[]{\textit{downregulates}} \mathit{Gene}\). Figure \ref{fig:disease-other-edge} shows the relationship between score and degree, as before, however here we stratified the genes by whether they have another type of relationship to the queried disease, for melanoma and Parkinson's Disease.

Curiously, while the pattern of high correlation between degree and score is maintained for both groups, there is little difference between them in terms of ranking. In other words, the existence of another relation type between the same disease and gene does not appear to affect the associates relation. Especially interesting are the genes at the lower end of the degree distribution, where there are genes whose only edge is to the disease of interest which are still ranked lowly. We could not replicate this with the other two, sparsely-connected, diseases since they did not have multiple relation types between them and genes.

\subsection{Graph Perturbations}
\label{ssec:graph-perturbations}

In this section we aim to understand whether we can artificially influence the rank assigned to a gene by a model via a perturbing process whereby edges are removed or added, thus altering the graph topology. Any change in score relative to a disease could then be measured by retraining the model on this new graph. For all experiments presented in this section, 10 repeats are performed over different sets of random edges, with results being presented as the mean with 95\% confidence intervals.

\textbf{Edge Removal.} In this experiment, a random selection edges are incrementally removed from the highest ranked novel gene for a given disease and any change in rank is measured. Two strategies are used for this removal process:
\begin{itemize}
	\item \emph{Disease:} edges are removed randomly from the gene to only other disease entities.
	\item \emph{Random:} edges are removed randomly from the gene to any other entity.
\end{itemize}

\begin{figure*}[!ht]
	\centering
	\begin{subfigure}[b]{0.48\textwidth}
		\centering
		\includegraphics[width=0.99\textwidth]{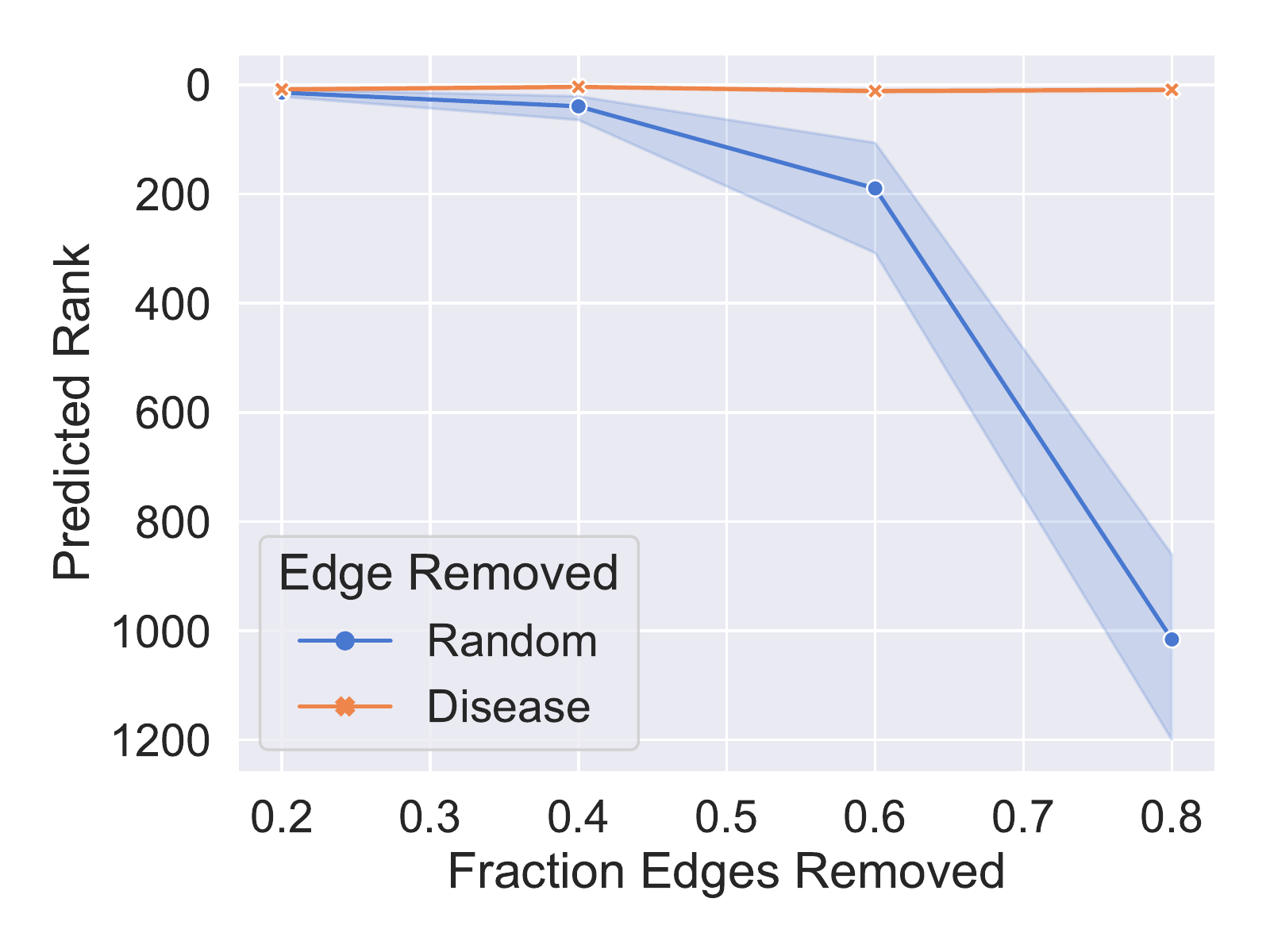}
		\caption{Leptin - Fallopian Tube Cancer}\label{fig:reduce:lep}
	\end{subfigure}
	\begin{subfigure}[b]{0.48\textwidth}
		\centering
		\includegraphics[width=0.99\textwidth]{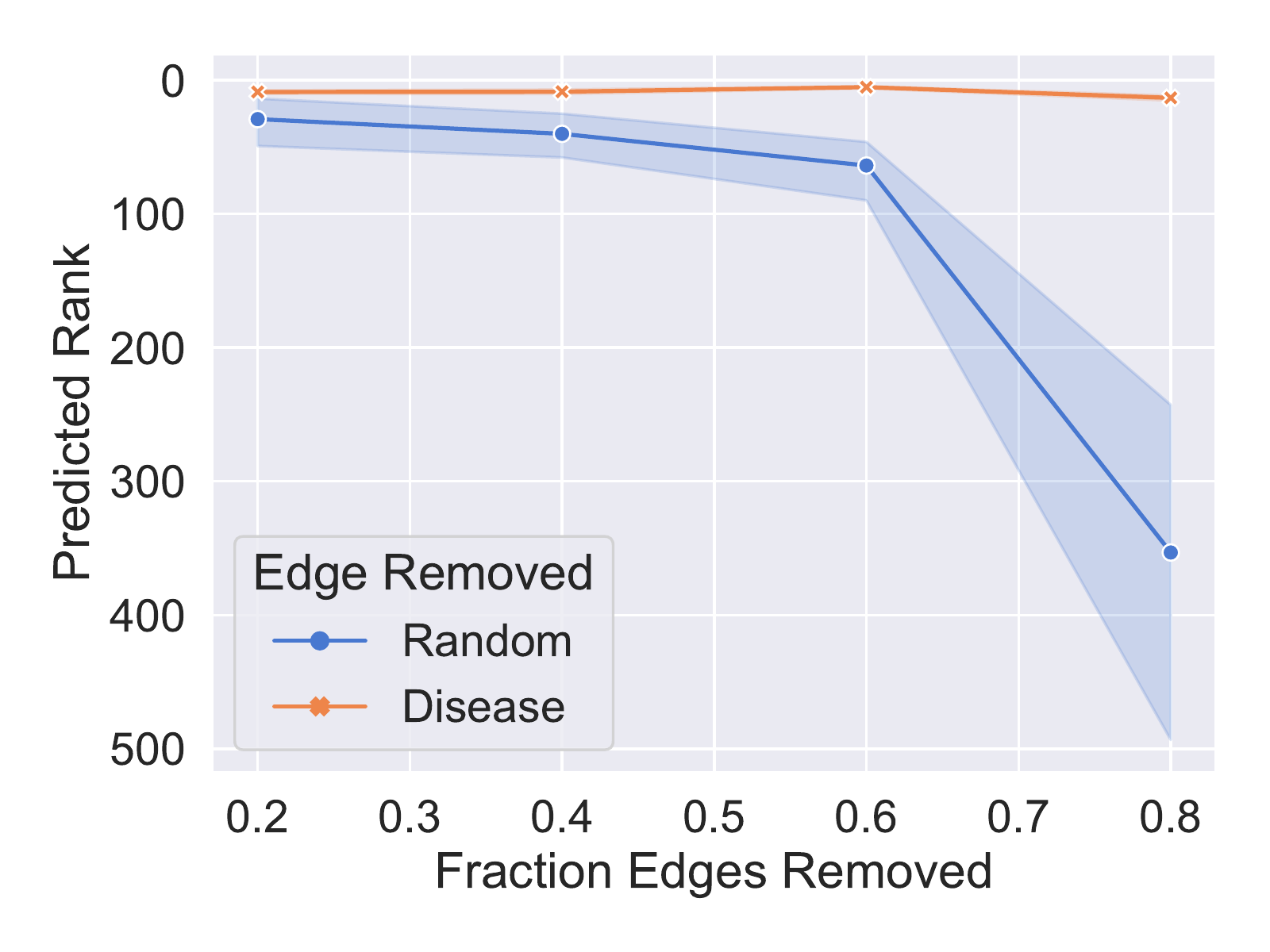}
		\caption{Interleukin 1 beta - Melanoma}\label{fig:reduce:inter}
	\end{subfigure}
	\caption{Removel of edges from the top ranked novel gene for two diseases: Fallopian Tube Cancer and Melanoma}
	\label{fig:reduce}
\end{figure*}

Figure \ref{fig:reduce} shows how the rank of the top gene for a disease is affected as different proportions of its edges are removed. As can be seen in both panels, there is a clear relationship between the removal of edges and a decrease in rank. It should noted that this result should not be considered surprising, indeed the assumption would be that as knowledge is removed from the graph, the model is able to learn less about the gene and thus is able to make less confident predictions about it. What is more interesting however is that by removing disease connections, the rank is relatively unaffected, which is somewhat counter-intuitive.

\textbf{Edge Addition.} In this experimental setup we add random, biologically meaningless edges to a gene to investigate how this impacts the score assigned to it by the model in relation to a set of diseases. To achieve this, we took the gene predicted to be the least likely to associate with a given disease and artificially increased its connectivity within the graph using two strategies:

\begin{itemize}
	\item \emph{Disease:} edges are added randomly from the gene to only other disease entities, excluding the disease we are scoring against. These edges conform to the graph schema and use the correct relation types of DaG, DuG and DdG.
	\item \emph{Ant-Comp-Gene:} edges are added randomly from the gene to other gene, compound and anatomy entities. Again these edges conform to graph schema and use the correct relation types.
\end{itemize}

\begin{figure*}[!ht]
	\centering
	\begin{subfigure}[b]{0.48\textwidth}
		\centering
		\includegraphics[width=0.99\textwidth]{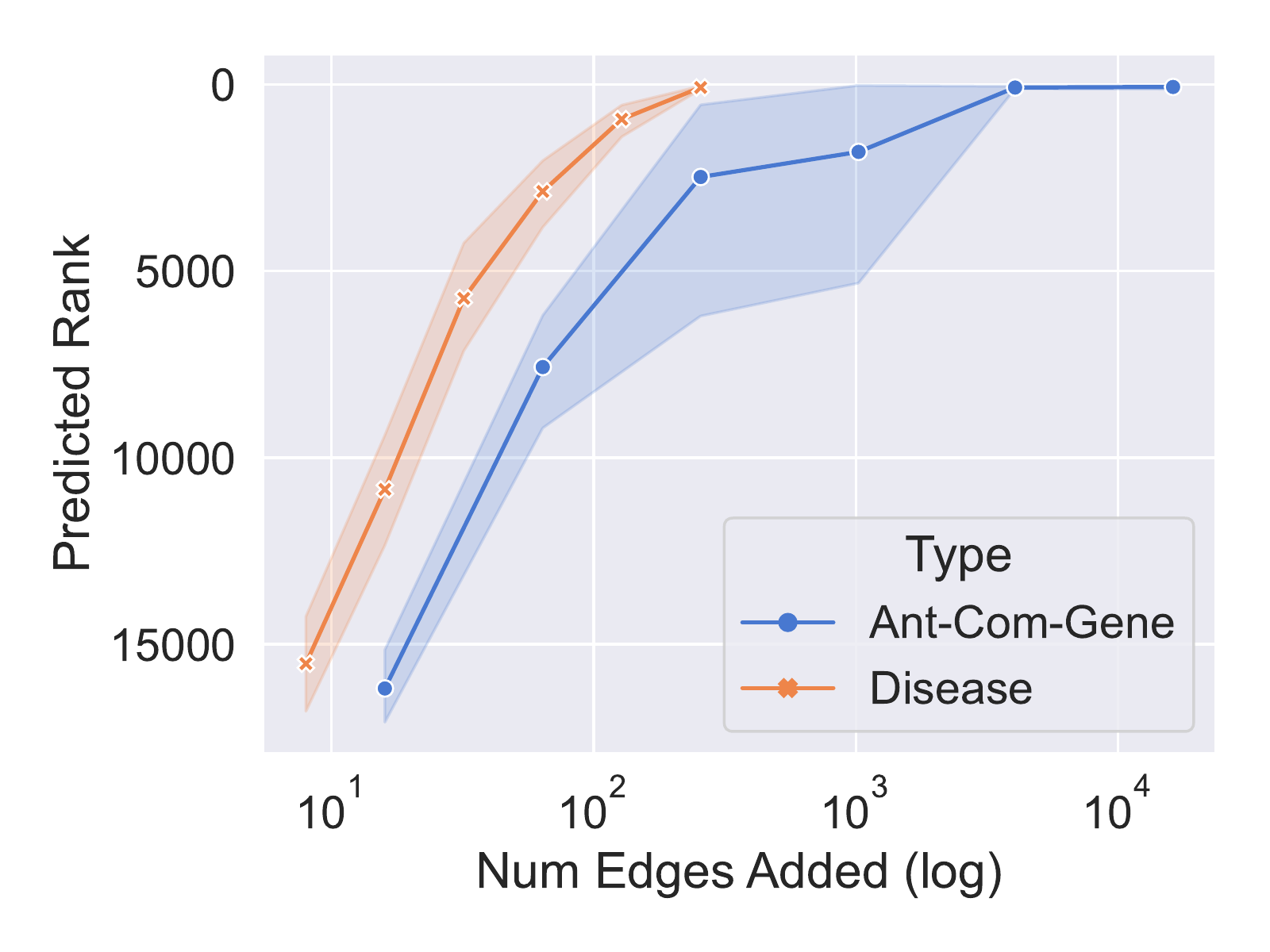}
		\caption{Fuchs endothelial dystrophy}\label{fig:add:fed}
	\end{subfigure}
	\begin{subfigure}[b]{0.48\textwidth}
		\centering
		\includegraphics[width=0.99\textwidth]{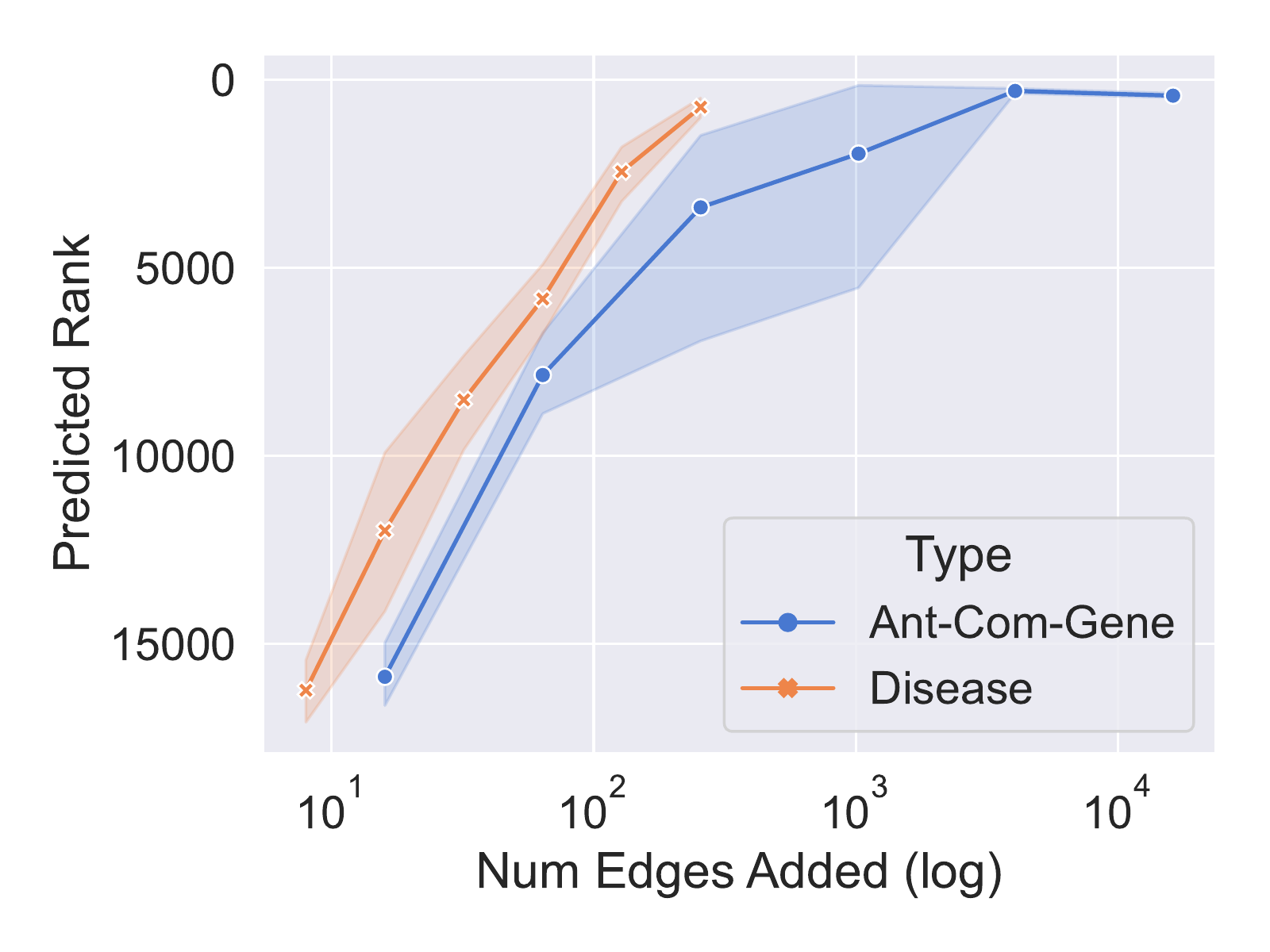}
		\caption{Breast Cancer}\label{fig:add:breast}
	\end{subfigure}
	\caption{Addition of edges the lowest ranked gene for two diseases: Breast Cancer and Fuchs endothelial dystrophy. In both cases, the gene TRG is used as the target for the edge addition.}
	\label{fig:add}
\end{figure*}

Figure \ref{fig:add:fed} shows the results for Fuchs endothelial dystrophy, whilst Figure \ref{fig:add:breast} the results for Breast Cancer. For both of these the gene to which the edges were added was T cell receptor gamma locus (TRG) as it was the least likely genes for both diseases. TRG has just a single edge in the original graph, connecting it to another gene. Overall the figures highlight the key observation that for both diseases, TRG was able to be moved from the least likely to one of the most likely gene to associate with the disease by the addition of biologically meaningless random edges. This finding provides further evidence that the predictions from KGE models can be biased by the frequency with which the entity is observed during the training process, and that this bias might outweigh any domain-specific knowledge captured in the data that may be useful to make more accurate real-world predictions.

\subsection{Case Study: UBC}
\label{ssec:ubc}

A consistent feature of the ranking vs degree plots is a gene to the top-right of the plot area, ranked curiously high no matter the disease in question, likely due to its particularly high degree. This gene is \emph{Polyubiquitin-C} (UBC), with a degree just below \(10^4\), nearly an order of magnitude higher than any other gene within Hetionet.

The UBC gene codes for multiple copies of the Ubiquitin protein, which in turn is a crucial component in many signalling and transport processes within the cell~\cite{kimura2010ubi}. Specifically, ubiquitin signalling is associated with protein degradation and recycling of amino acids, which is a critical metabolic process that has important implications for normal function of a cell. With such an critical role, it is not surprising that ubiquitin, and thus indirectly ubiquitin-coding genes like UBC, are involved in many pathways.

Additionally, the overwhelming majority of the connections UBC has in Hetionet is to other genes (Table \ref{tab:ubc-conn}), likely due to UBC being directly responsible for degradation of many proteins these genes code for. Despite its predicted association with all our diseases of interest, UBC has a only single connection to a Disease and Compound entity type in the graph.

\begin{table}[ht!]
	\centering
	\begin{tabular}{l c c}
		\toprule
		\textbf{Neighbour Type} & \textbf{\# of Conns} & \textbf{\# of Distinct} \\ \midrule \midrule
		Gene                    & 8789                 & 8653                    \\
		BiologicalProcess       & 345                  & 345                     \\
		Pathway                 & 173                  & 173                     \\
		Anatomy                 & 55                   & 44                      \\
		CellularComponent       & 7                    & 7                       \\
		MolecularFunction       & 1                    & 1                       \\
		Compound                & 1                    & 1                       \\
		Disease                 & 1                    & 1                       \\ \bottomrule
	\end{tabular}
	\vspace{5pt}
	\caption{Overview of UBC's connectivity within Hetionet. The disparity between number of connections and the number of distinct neighbours is due to the fact that any two entities can have multiple relationships between them.}
	\label{tab:ubc-conn}
\end{table}

Seen altogether, UBC provides a good example of where a highly connected entity is being over ranked as a prediction using KGE methods. UBC is ranked in the top 100 for every disease in the Hetionet dataset (See Table \ref{tab:ubc-rank}), despite having only a single relationship of the desired type (i.e. DaG), or barely any type of relationship to the entity type of interest (i.e. Disease). Furthermore, a gene that is heavily interconnected with other genes, like UBC, is likely to be considered a poor candidate disease target since any interference with its function will likely have many unintentional and undesired downstream effects, as vital metabolic and signalling processes would be altered out of the context of the disease.

As an interesting point of comparison, in the OpenBioLink dataset~\cite{breit2020openbiolink}, UBC has a degree of 5937 and is only the sixth most well-connected gene, with many genes possessing a similar degree value. Table \ref{tab:ubc-rank} shows how in OpenBioLink, now that UBC is no longer an outlier with nearly an order of magnitude greater connections than the next gene, its predicted rank in relation to the diseases is greatly reduced.

\begin{table}[ht!]
	\centering
	\begin{tabular}{l c c}
		\toprule
		\textbf{Disease}            & \textbf{Rank (Hetionet)} & \textbf{Rank (OpenBioLink)} \\ \midrule \midrule
		Breast Cancer               & 60                       & 232                         \\
		Melanoma                    & 82                       & 1061                        \\
		Parkinsons disease          & 59                       & 1599                        \\
		Fuchs Endothelial Dystrophy & 23                       & 3329                        \\
		Fallopian Tube Cancer       & 52                       & 2117                        \\
		\bottomrule
	\end{tabular}
	\vspace{5pt}
	\caption{The predicted rank of the gene UBC to be associated with the range of diseases. Results presented using the Hetionet and OpenBioLink graphs with the TransE model.}
	\label{tab:ubc-rank}
\end{table}

\textbf{Graph Rewiring.} Using UBC as our gene of interest, we investigated one final way of altering the graph topology: random rewiring. Here, edges from UBC to other entities in the graph are permuted by swapping the target entity via a random strategy. In this strategy, edges are rewired to the correct entity type. Further, the existing set of entities were removed from the pool of possible replacements to avoid the new edge replicating an already observed edge in the training data.

\begin{figure*}[!ht]
	\centering
	\begin{subfigure}[b]{0.48\textwidth}
		\centering
		\includegraphics[width=0.99\textwidth]{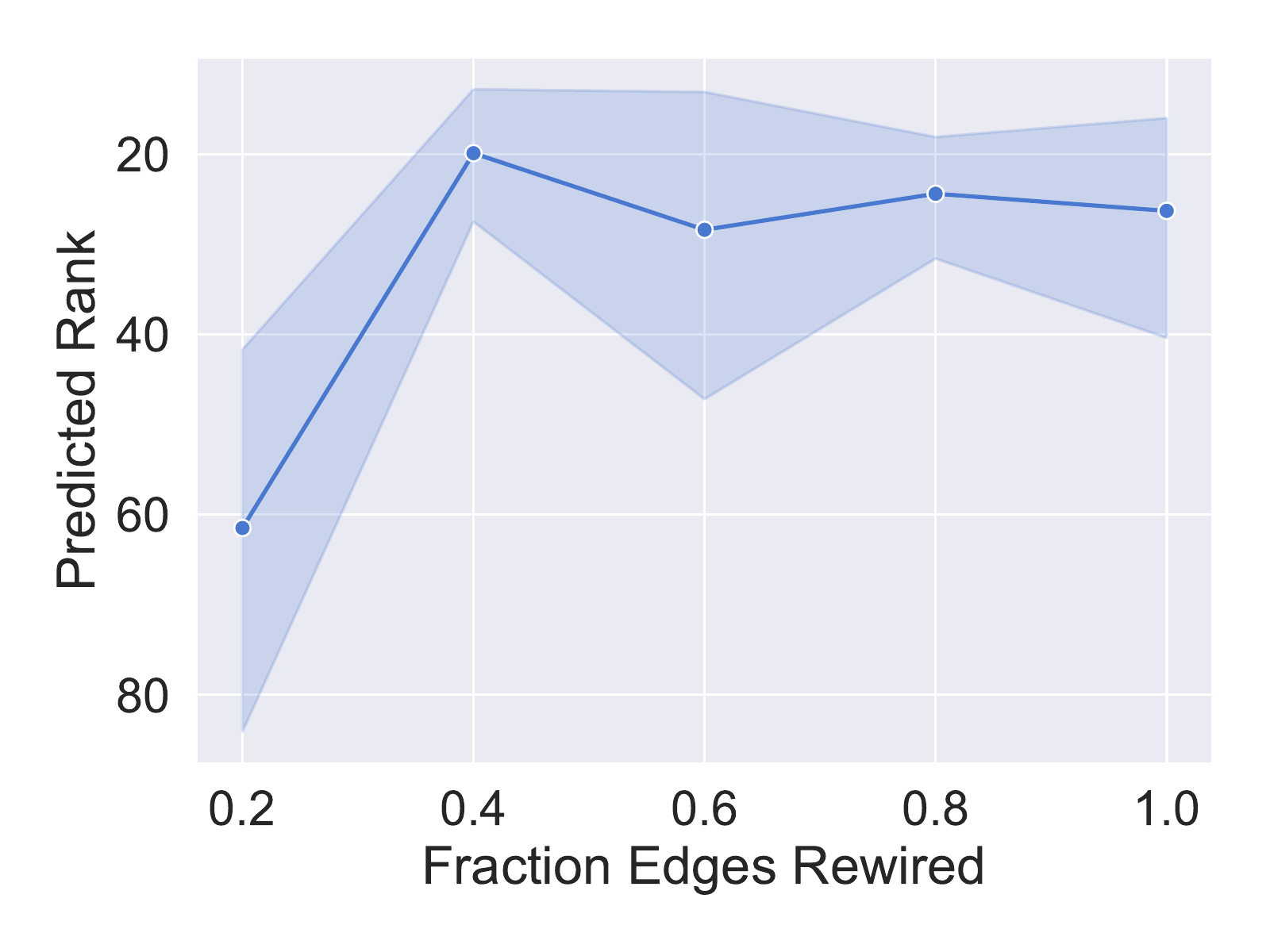}
		\caption{Melanoma}\label{fig:rewire:mel}
	\end{subfigure}
	\begin{subfigure}[b]{0.48\textwidth}
		\centering
		\includegraphics[width=0.99\textwidth]{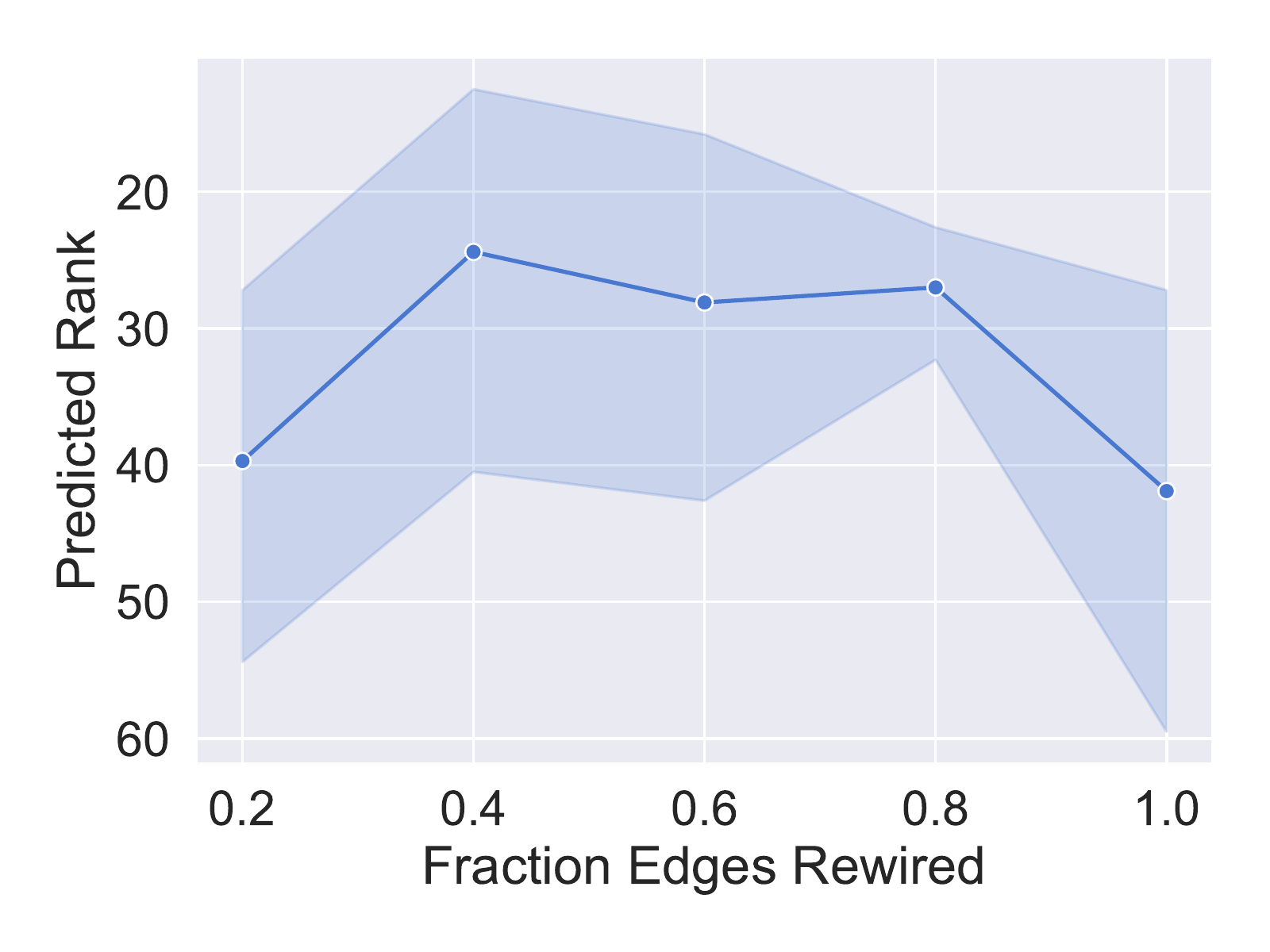}
		\caption{Parkinsons}\label{fig:rewire:park}
	\end{subfigure}

	\begin{subfigure}[b]{0.48\textwidth}
		\centering
		\includegraphics[width=0.99\textwidth]{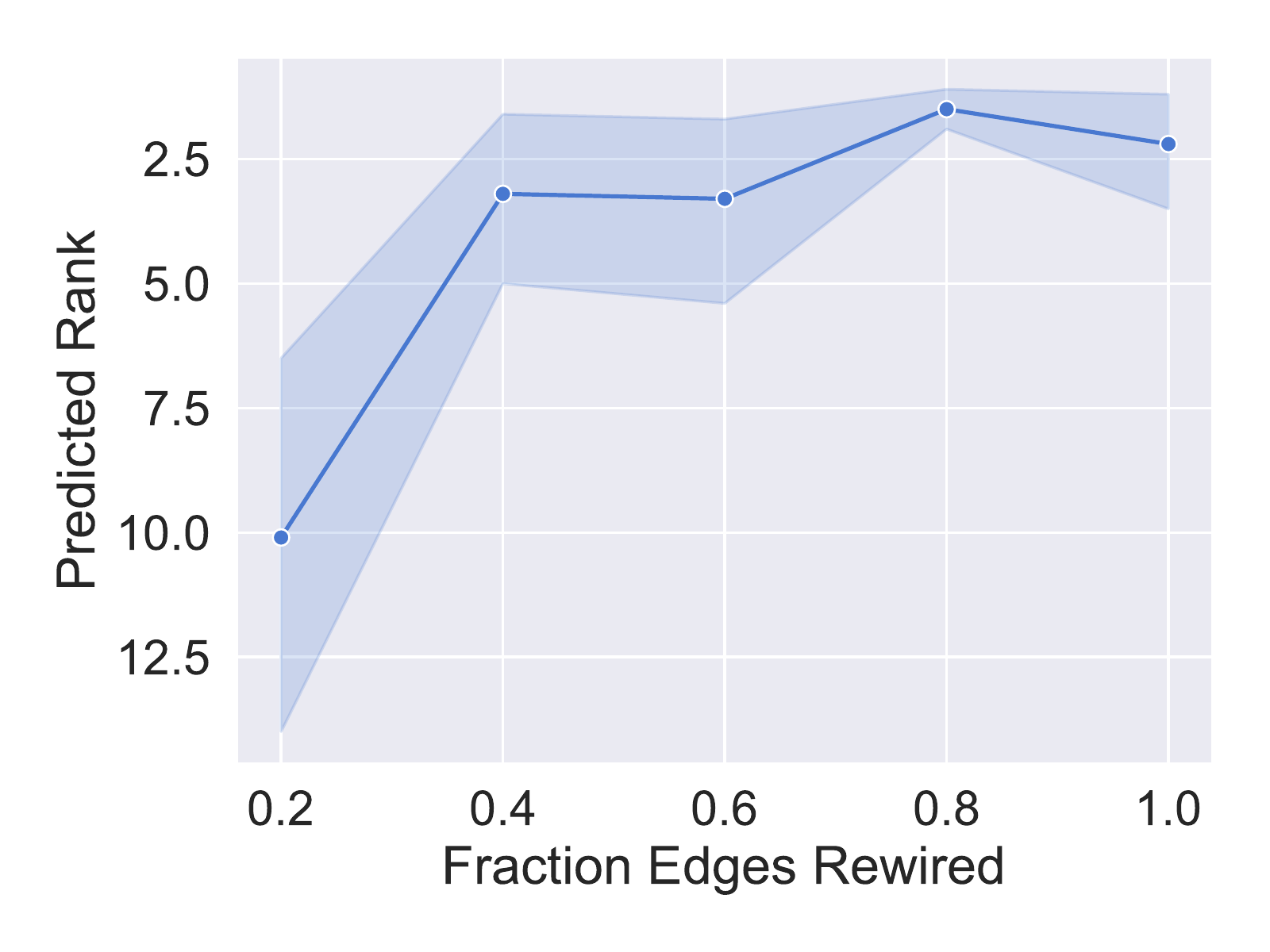}
		\caption{Fuchs Endothelial Dystrophy}\label{fig:rewire:fuchs}
	\end{subfigure}
	\begin{subfigure}[b]{0.48\textwidth}
		\centering
		\includegraphics[width=0.99\textwidth]{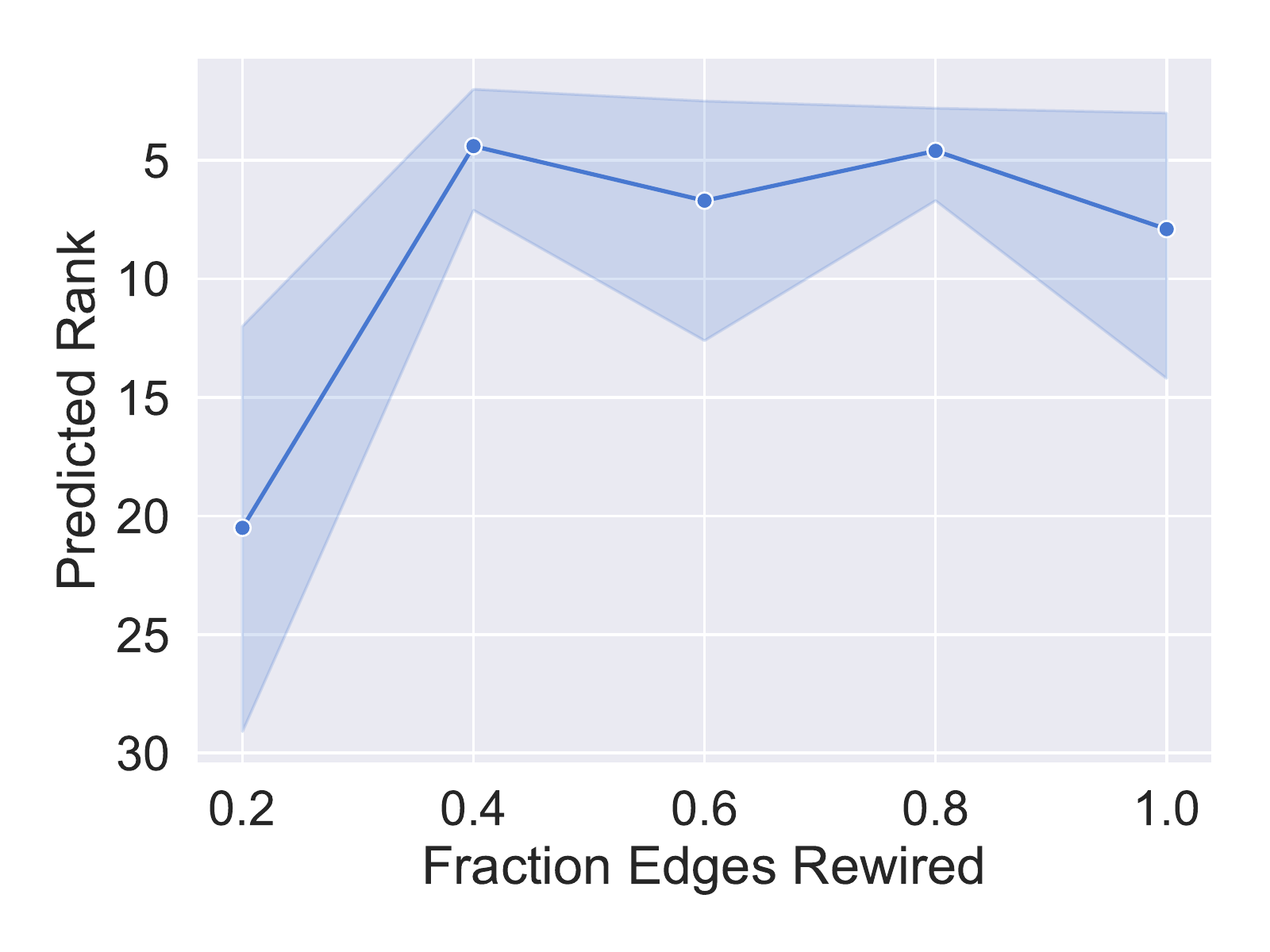}
		\caption{Fallopian Tube Cancer}\label{fig:rewire:fal}
	\end{subfigure}
	\caption{The change in rank assigned to the gene UBC after a varying fraction of its edges are rewired.}
	\label{fig:rewire}
\end{figure*}

Figure \ref{fig:rewire} demonstrates the results of this rewiring process across four diseases in relation to UBC. The most apparent observation is that rewiring edges in accordance to the schema has almost no negative impact on the rank of UBC in relation to the four diseases. Indeed the rank actually increases for UBC across all diseases as we rewire edges. We hypothesise this to be caused by new edges being formed with other hub entities, thus further increasing UBC's importance within the graph. Simply put, this experiment shows we are able to go as far as completely randomising the biological information contained within UBC's edges without drastically altering how likely the model considers the gene to be associated with a disease. This is further evidence that KGE models are seemingly biased by the number of connections an entity has, rather than any domain knowledge encoded within its edges.

\subsection{Discussion}

The purpose of this study was not to critique a particular dataset or methodology, the trends we observed were present across multiples of both anyway, rather we wanted to take a deeper look at how topological imbalance may be affecting predictions within the drug discovery domain. We believe the evidence from this experimental evaluation suggests that degree has a strong influence on KGE models, with highly connected entities seemingly being highly ranked regardless of context. Whilst this result may not initially seem surprising, as the high-degree entities are simply observed more during the model training process, the level of correlation should nevertheless be of concern to practitioners.

Indeed the collective evidence from this study raises the prospect that domain knowledge encoded in the relationships matters less than one would expect and a highly connected, but biologically implausible, gene can be associated with a disease over one that would ultimately make a better drug target. This result also suggests that the predictions made by a KGE model, in the context of target discovery, may suffer from non-specificity -- the same set of highly connected genes are highly ranked regardless of the query disease, leading to little variability in candidates.

\textbf{Modeling.} Many existing KGE models were designed to solve tasks in other domains, and may not have been developed against biomedical graphs which can often have a distinct topological structure, such as higher average connectivity~\cite{liu2021neural}. We hope that new models will more regularly include datasets such as Hetionet during their development phase. More generally, and taking cues from the field of GNNs~\cite{liu2021tail, liu2020towards, tang2020investigating}, new methods could be be developed which consider how best to learn meaningful representations for low-degree entities.

\textbf{Evaluation.} Common metrics used for evaluating KGE models such as Hits@k and MRR do not consider entity or relation frequency and thus can be biased by it~\cite{mohamed2020popularity}. As a simple mitigating step, care should be taken to ensure that low degree entities are included in test and validation sets to ascertain their impact is reflected in these metrics. Overall,  global performance metrics should not be the sole way of assessing model performance, rather practitioners should directly inspect the output predictions to look for any patterns contained within. For example, a highly ranked, but low degree, entity may be particularly interesting to consider.

\textbf{Graph Composition.} The examples demonstrated here provide a cautionary tale as relationship types seem to matter less than the overall volume of connectivity. This implies that KG practitioners should be aware of the implications of their data modelling decisions, being cognisant of any potential outliers in terms of topological structure. Thought should also be given to whether auxiliary information, which may un-proportionally increase the degree of certain entities, is really beneficial to the task at hand. This is especially important if edges are being automatically extracted, from publications for example, via NLP-based pipelines. Depending on the biological question at hand, a more focused and task-specific graph projection may lead to better overall predictions~\cite{ratajczak2021task}.

Throughout this process we have determined the following initial set of recommendations to help address topological imbalance:

\begin{itemize}
	\item \emph{Awareness -} We encourage practitioners to actively investigate the topologies of the graphs they are using for predictions, checking specifically for any inherent imbalance in entity or relation frequency. This also extends to awareness of which data sources and modalities have been used in graph construction, and any redundancies that may lie within.
	\item \emph{Predictions -} Further, KGE users should not rely solely on performance metrics, instead the ranked list of predictions should be inspected and compared with topological features for those entities. Where metrics are required, results should be presented across different levels of connectivity.
	\item \emph{Projections -} Practitioners should consider creating task-specific graph projections from larger holistic resources like Hetionet which may reduce the overall volume of connections and thus imbalance. However this should be performed in a principled manner so pertinent information is not lost as the graph is reduced.
	\item \emph{Edge Confidence -} One way to explore more principled projections could be to filter the graph based on confidence scores assigned to the edges. These values could be taken directly from the underlying data sources or computed where absent.
\end{itemize}
\section{Conclusion}\label{sec:conclusion}

Biological knowledge graphs, especially in a drug discovery context, are heavily unbalanced in terms of degree distribution. On the one hand we observe prior knowledge bias, i.e.\ well-studied genes and diseases are typically have more, and often more confident, annotations. On the other hand, regardless of any \emph{a priori} bias, biological entities are known to display a power law of connectivity, some genes have a wide range of biological functions, involved in various different diseases, and typically work in concert with a wide range of other genes. These two factors combine to create super-hub entities in the graph.

Additionally, modelling choices during the conception of the KG may lead to a very connection-dense resultant graph. For example, in the case of Hetionet, almost any two gene entities can be connected with a 1-hop path via an anatomy or cellular-component entity. This complexity is increased when one considers the multitude of ways entities can be connected to each other, depending on the data sources integrated into the graph. Automated data mining and natural text processing pipelines may further impact this by adding a significant number of relationships between the entities, often with questionable certainty, or little contextual information.

In this study, we demonstrate a reproducible tendency of KGE methods to overestimate highly-connected entities in predictive tasks. Furthermore, we show that this particular issue is independent of the choice of entities, or indeed even type of predictive task. We discuss the potential reasons for these results, as well as propose some tangible solutions that can be applied to avoid or at least ameliorate these type of problems. In the light of these findings, we believe it is prudent to pay close attention to the data modeling when creating a biological KG, application of embedding methods as well as to validate the resultant rankings for predictive tasks on these KGs. Despite the shortcomings discussed in this paper, we believe KGs and KGE methods provide valuable tools for generating meaningful representations of biological processes, reducing machine learning complexity and thus have potential for real impact in drug discovery efforts, especially when used appropriately for the underlying domain specific data. For future work, we aim to reduce the impact of connectivity imbalance through the use of a popularity bias term introduced in the model interaction function.

\section*{Acknowledgement}

We would like to thank all of the PyKEEN team for their help and support. We would also like to acknowledge the use of the Science Compute Platform (SCP) within AstraZeneca. Stephen Bonner is a fellow of the AstraZeneca postdoctoral program.

\bibliographystyle{plain}
\bibliography{RPbib}

\end{document}